\documentclass[twoside,11pt]{article}

\usepackage{jmlr2e2}

\usepackage{epstopdf}
\usepackage{amsmath, mathabx}
\usepackage{mathrsfs}
\usepackage{hyperref}
\usepackage{comment}
\usepackage{xcolor}
\usepackage{enumitem}
\usepackage{algorithm}
\usepackage{algpseudocode}
\usepackage{stmaryrd}
\usepackage{array}
\usepackage[utf8]{inputenc} 
\usepackage[T1]{fontenc}
\usepackage{fancyhdr}
\usepackage{multirow}
\usepackage{subfigure}
\newcommand{\0}{{\mathbf{0}}}

\renewcommand{\v}{{\mathbf{v}}}
\newcommand{\e}{{\mathbf{e}}}
\renewcommand{\u}{{\mathbf{u}}}
\newcommand{\scA}{\mathscr{A}}
\renewcommand{\a}{{\mathbf{a}}}
\newcommand{\y}{{\mathbf{y}}}

\newcommand{\B}{{\mathbf{B}}}

\newcommand{\D}{{\mathbf{D}}}
\newcommand{\R}{{\mathbf{R}}}
\newcommand{\U}{{\mathbf{U}}}
\newcommand{\V}{{\mathbf{V}}}
\newcommand{\W}{{\mathbf{W}}}

\newcommand{\I}{{\mathbf{I}}}

\newcommand{\M}{{\mathbf{M}}}

\newcommand{\sA}{{\mathscr{A}}}

\newcommand{\bcX}{{\mathbfcal{X}}}
\newcommand{\bcA}{{\mathbfcal{A}}}
\newcommand{\bcY}{{\mathbfcal{Y}}}

\newcommand{\bcZ}{{\mathbfcal{Z}}}
\newcommand{\bcB}{{\mathbfcal{B}}}
\newcommand{\bcS}{{\mathbfcal{S}}}

\newcommand{\bcE}{{\mathbfcal{E}}}
\newcommand{\bcI}{{\mathbfcal{I}}}
\newcommand{\bcV}{{\mathbfcal{V}}}

\newcommand{\br}{\mathbf{r}}

\newcommand{\cH}{\mathcal{H}}

\newcommand{\cL}{\mathcal{L}}

\newcommand{\cM}{\mathcal{M}}

\newcommand{\cT}{\mathcal{T}}

\newcommand{\bbR}{\mathbb{R}}

\newcommand{\bbO}{\mathbb{O}}
\newcommand{\bbM}{\mathbb{M}}

\newcommand{\bvarepsilon}{\boldsymbol{\varepsilon}}

\newcommand{\rmvec}{{\rm{vec}}}
\renewcommand{\exp}{{\rm{exp}}}

\newcommand{\SVD}{{\rm{SVD}}}

\newcommand{\argmin}{\mathop{\rm arg\min}}
\newcommand{\argmax}{\mathop{\rm arg\max}}
\newcommand{\rank}{{\rm rank}}
\newcommand{\tuckerrank}{{\rm Tucrank}}
\newcommand{\tHS}{{\rm HS}}
\newcommand{\grad}{{\rm grad \,}}
\newcommand{\Range}{{\mathcal{R}}}
\newcommand{\QR}{{\rm QR}}

\newcommand{\Proj}{P}

\newtheorem{Definition}{Definition}
\newtheorem{Theorem}{Theorem}

\newtheorem{Lemma}{Lemma}
\newtheorem{Remark}{Remark}
\newtheorem{Corollary}{Corollary}

\newtheorem{Proposition}{Proposition}

\DeclareMathAlphabet\mathbfcal{OMS}{cmsy}{b}{n}

\usepackage{lastpage}
\jmlrheading{24}{2023}{1-\pageref{LastPage}}{4/21; Revised
11/22}{7/23}{21-0438}{Yuetian Luo and Anru R. Zhang}
\ShortHeadings{title}{Luo and Zhang}

\ShortHeadings{Second-order Method for Low-rank Tensor Estimation}{Y. Luo and A. R. Zhang}
\firstpageno{1}

\begin{document}

\title{Low-rank Tensor Estimation via Riemannian Gauss-Newton: Statistical Optimality and Second-Order Convergence}

\author{\name Yuetian Luo \email yuetian@uchicago.edu \\
       \addr Data Science Institute\\
       University of Chicago\\
       Chicago, IL 60637, USA
       \AND
       \name Anru R. Zhang \email anru.zhang@duke.edu \\
       \addr Departments of Biostatistics \& Bioinformatics and Computer Science\\
       Duke University\\
       Durham, NC 27710, USA}
\editor{Suvrit Sra}

\maketitle

\begin{abstract}
    In this paper, we consider the estimation of a low Tucker rank tensor from a number of noisy linear measurements. The general problem covers many specific examples arising from applications, including tensor regression, tensor completion, and tensor PCA/SVD. We consider an efficient Riemannian Gauss-Newton (RGN) method for low Tucker rank tensor estimation. Different from the generic (super)linear convergence guarantee of RGN in the literature, we prove the first local quadratic convergence guarantee of RGN for low-rank tensor estimation in the noisy setting under some regularity conditions and provide the corresponding estimation error upper bounds. A deterministic estimation error lower bound, which matches the upper bound, is provided that demonstrates the statistical optimality of RGN. The merit of RGN is illustrated through two machine learning applications: tensor regression and tensor SVD. Finally, we provide the simulation results to corroborate our theoretical findings.
\end{abstract}


\begin{sloppypar}
\section{Introduction} \label{sec: introduction}

The past decades have seen a large body of work on tensors or multiway arrays in applied mathematics, signal processing, machine learning, statistics, among many other fields. Tensors arise in numerous applications involving multiway data, such as brain imaging \citep{zhou2013tensor,zhang2019tensor}, electron microscopy imaging \citep{han2020optimal,zhang2020denoising}, recommender system design \citep{bi2018multilayer}. In addition, tensor methods have been applied to many problems in statistics and machine learning where the observations are not necessarily tensors, such as topic and latent variable models \citep{anandkumar2014tensor}, additive index models \citep{balasubramanian2018tensor}, high-order interaction pursuit \citep{hao2018sparse}. In this paper, we focus on a prototypical model for tensor estimation: 
\begin{equation}\label{eq:model}
\y = \scA (\bcX^*) + \bvarepsilon.
\end{equation}
Here, $\y, \bvarepsilon \in\mathbb{R}^n$ are the observations and unknown noise and $\bcX^* \in \bbR^{p_1 \times \cdots \times p_d}$ is an order-$d$ parameter tensor of interest. $\scA: \mathbb{R}^{p_1\times \cdots \times p_d} \to \mathbb{R}^n$ is a known linear map, which can be explicitly expressed as
\begin{equation} \label{eq: affine operator}
\scA(\bcX^*) = \left[\langle\bcA_1, \bcX^*\rangle, \ldots, \langle\bcA_n, \bcX^*\rangle \right]^\top, \langle \bcA_i, \bcX^*\rangle = \sum_{1\leq i_k\leq p_k, 1\leq k\leq d}(\bcA_i)_{[i_1,\ldots,i_d]} \bcX^*_{[i_1,\ldots,i_d]}
\end{equation}
with the given measurement tensors $\{\bcA_i\}_{i=1}^n \subseteq \mathbb{R}^{p_1\times \cdots \times p_d}$. Our goal is to estimate $\bcX^*$ based on $(\y, \scA)$. When $\bvarepsilon = 0$, \eqref{eq:model} becomes the low-rank tensor recovery problem \citep{rauhut2017low} where the aim is to recover $\bcX^*$ exactly. 

In many applications, $\prod_{k=1}^d p_k$, i.e., the number of parameters in $\bcX^*$, is much greater than the sample size $n$, so some structural conditions are often assumed to ensure the problem is well-posed. In the literature, the low-rank assumption was widely considered \citep{kolda2009tensor,zhou2013tensor,anandkumar2014guaranteed,richard2014statistical,montanari2018spectral}. In this work, we focus on the setting that the target parameter $\bcX^*$ is low Tucker rank and admits the following Tucker (or multilinear) decomposition with Tucker rank $\br = (r_1,\ldots, r_d)$:
\begin{equation}\label{eq: decomposition of X}
	\bcX^* = \bcS \times_{1} \U_1 \times \cdots \times_{d} \U_d.
\end{equation}
Here, $\bcS \in \bbR^{r_1 \times \cdots \times r_d}$ is the order-$d$ core tensor; $\U_k$ is a $p_k$-by-$r_k$ matrix with orthonormal columns, which represents the mode-$k$ singular vectors of $\bcX^*$; ``$\times_{k}$'' is the tensor-matrix product along mode $k$. The formal definitions of Tucker decomposition and tensor-matrix product are given in Section \ref{sec:notations}. 

With different designs of $\scA$, the general model~\eqref{eq:model} covers many specific settings arising from applications, such as recommender system \citep{bi2018multilayer}, neuroimaging \citep{guhaniyogi2015bayesian,li2017parsimonious}, longitudinal relational data analysis \citep{hoff2015multilinear}, imaging processing \citep{guo2012tensor}. The specific settings of model \eqref{eq:model} include: 
\begin{itemize}[leftmargin=*]
	\item {\bf Tensor regression with general random or deterministic design} \citep{zhou2013tensor,raskutti2015convex}, where $\bcA_i$ are general tensors. Specifically, the Gaussian ensemble design ($\bcA_i$ has i.i.d. Gaussian/sub-Gaussian entries) is widely studied in the literature.
    \item {\bf Tensor completion} \citep{gandy2011tensor,liu2013tensor,yuan2014tensor}: $\bcA_i = \e_{a^{(i)}_1} \circ \cdots \circ \e_{a^{(i)}_d}$, $\e_{a_k^{(i)}} $ is the $a_k^{(i)}$th canonical vector and $\{a^{(i)}_1, \cdots, a^{(i)}_d\}_{i=1}^n$ are randomly selected integers from $[p_1]\times \cdots \times [p_d]$, ``$\circ$'' represents the outer product and $[p_k] = \{1,\ldots, p_k\}$; 
    \item {\bf Tensor estimation via rank-1 projections} \citep{hao2018sparse}: $\bcA_i = \a_1^{(i)} \circ \cdots \circ \a_d^{(i)}$, where $\{\a_k^{(i)} \in \bbR^{p_k} \}_{k=1}^d$ are random vectors; 
    \item {\bf Tensor PCA/SVD} \citep{richard2014statistical,hopkins2015tensor,zhang2018tensor,perry2020statistical} is a special case of tensor completion where all entries are observable. In this particular setting, we can tensorize $\y, \bvarepsilon$ and rewrite the model \eqref{eq:model} equivalently to $\bcY = \bcX^* + \bcE$. Here $\bcX^*$ is the low Tucker rank signal tensor and $\bcE$ is the noise.
\end{itemize}

In view of model \eqref{eq:model} and assumption \eqref{eq: decomposition of X}, a natural estimator of $\bcX^*$ is
\begin{equation}\label{eq:minimization}
\widehat{\bcX} =  \argmin_{\bcX \in \mathbb{R}^{p_1\times \cdots \times p_d}} f(\bcX) := \frac{1}{2} \left\|\y - \scA(\bcX)\right\|_2^2, \quad \text{subject to}\quad \tuckerrank(\bcX)= \br.
\end{equation}
Here $\tuckerrank(\bcX)$ is the Tucker rank of $\bcX$ (see definition in Section \ref{sec:notations}). However, the optimization problem in \eqref{eq:minimization} is non-convex and NP-hard in general. To tame the non-convexity, a common scheme is the convex relaxation \citep{mu2014square,raskutti2015convex,tomioka2011statistical}. However, this scheme may either obtain suboptimal statistical guarantees or require evaluating the tensor nuclear norm, which is NP-hard to compute in general \citep{hillar2013most}. Alternatively, a large body of literature turns to the non-convex formulation and focuses on developing computationally efficient two-stage procedures for estimating $\bcX^*$: first, one obtains a warm initialization of $\bcX^*$ and then runs local algorithms to refine the estimate. Provable guarantees on estimation or recovery of $\bcX^*$ for such a two-stage paradigm have been developed in different scenarios \citep{rauhut2017low,chen2016non,ahmed2020tensor,han2020optimal,cai2019nonconvex,hao2018sparse,cai2020provable,xia2017statistically}. In this work, we focus on the non-convex formulation 
and aim to develop a provable computationally efficient estimator for $\bcX^*$. Departing from the existing literature that focuses on the first-order local methods, we consider a Riemannian Gauss-Newton (RGN) algorithm for iterative refinement and establish the first quadratic convergence guarantee on the estimation of $\bcX^*$.

\subsection{Our Contributions} \label{sec: contributions}

In this paper, we develop a new Riemannian Gauss-Newton (RGN) algorithm for low-rank tensor estimation. The proposed algorithm is tuning-free and generally has the same per-iteration computational complexity as the alternating minimization \citep{zhou2013tensor,li2013tucker} and comparable complexity to the other first-order methods including projected gradient descent \citep{chen2016non} and gradient descent \citep{han2020optimal}. 

\begin{figure}[h!]
  	\centering
  	\subfigure[Tensor regression. Here, $n$ is the sample size, $\bcX^* \in \bbR^{p \times p \times p}$ with $p = 30$, Tucker rank $\br = (3,3,3)$, $\bvarepsilon \overset{i.i.d}\sim N(0, \sigma^2)$ with $\sigma \in \{0,10^{-6} \}$ and $\bcA_i$ has i.i.d. standard Gaussian entries.]{\includegraphics[width=0.46\textwidth]{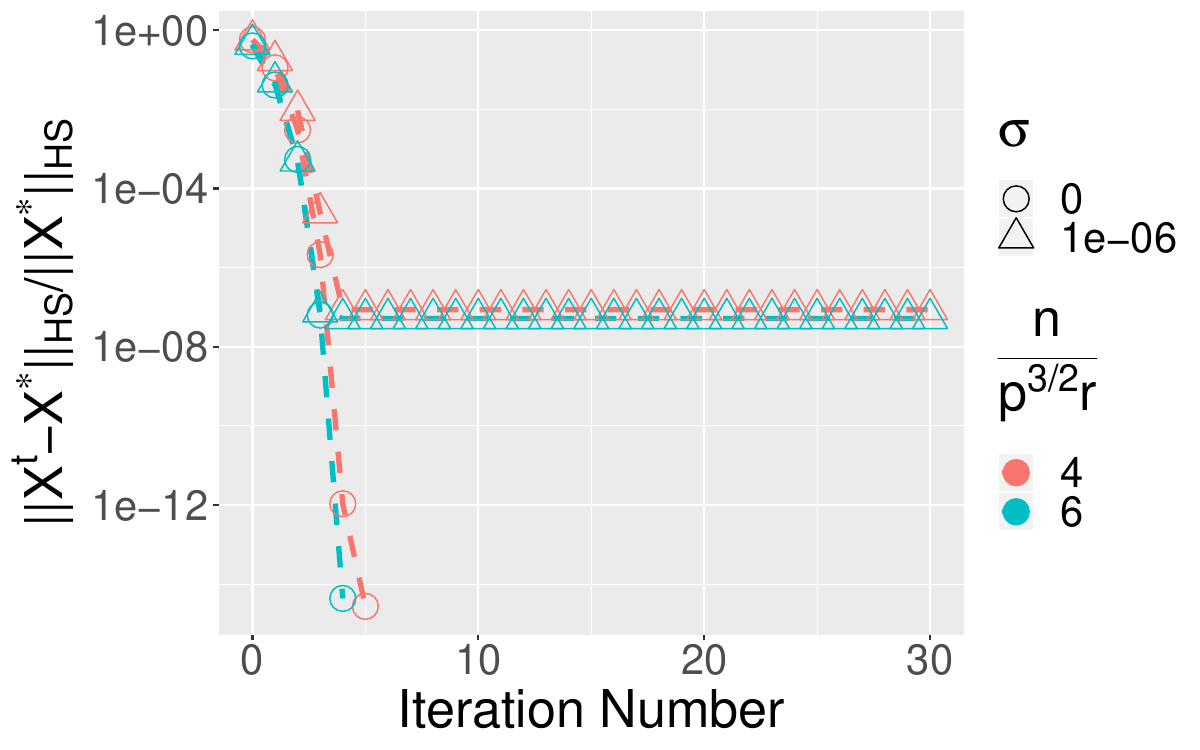}}
  	\qquad
  	\subfigure[Tensor completion. Here, we observe partial uniformly-at-random sampled entries from the noisy tensor $\bcY$ index by $\Omega$, where $\bcY = \bcX^* + \bcE \in \bbR^{p \times p \times p}$ with $p = 50$. Tucker rank of $\bcX^*$ is $(3,3,3)$, $n = |\Omega|$ and $\bcE$ has i.i.d. $N(0, \sigma^2)$ entries with $\sigma \in \{0,10^{-6} \}$.]{\includegraphics[width=0.46\textwidth]{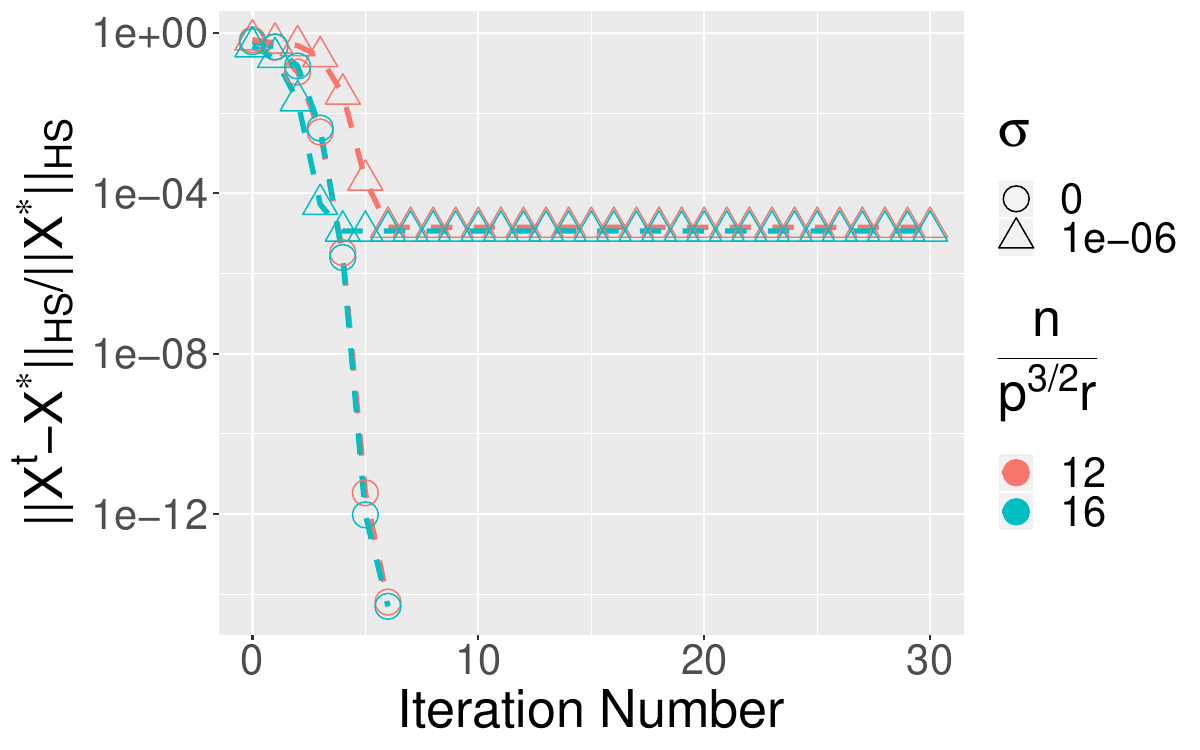}}
  	\caption{RGN achieves a quadratic rate of convergence in low-rank tensor estimation. More details of the simulation setting are given in Section \ref{sec:numerics}.}\label{fig: RGN performance illustration}
\end{figure}

Moreover, assuming $\scA$ satisfies the tensor restricted isometry property (TRIP) (see Definition \ref{def: RIP}), we prove that with some proper initialization, the iterates generated by RGN converge quadratically to $\bcX^*$ up to some statistical error. Especially in the noiseless setting, i.e., $\bvarepsilon = 0$, RGN converges quadratically to the exact parameter $\bcX^*$. Figure \ref{fig: RGN performance illustration} shows the numerical performance of RGN in tensor regression (left panel) and tensor completion (right panel): in the noiseless case, RGN converges quadratically to $\bcX^*$; in the noisy case, RGN converges quadratically to a neighborhood of $\bcX^*$ up to some statistical error. More simulation results on tensor estimation via rank-1 projections and tensor SVD can be found in Section \ref{sec:numerics}. 
Since RGN generally converges to a point with nonzero function value in the noisy setting, the generic theory on RGN can only guarantee a (super)linear convergence rate to a stationary point \citep{absil2009optimization,breiding2018convergence}. Our result complements the classic theory of RGN: we show RGN converges quadratically to a neighborhood of the true parameter of interest, which achieves a statistically optimal estimation error rate. To our best knowledge, such a result is new and our RGN is the first algorithm with a provable guarantee of second-order convergence for the low-rank tensor estimation. 

Furthermore, we provide a deterministic minimax lower bound for the estimation error under model \eqref{eq:model}. The lower bound matches the estimation error upper bound, which demonstrates the statistical rate-optimality of RGN. 

Next, we apply RGN to two problems arising from applications in machine learning and statistics: tensor regression and tensor SVD. In both problems, we prove the iterates of RGN converge quadratically to a neighborhood of $\bcX^*$ that achieves the minimax optimal estimation error. A comparison of RGN and prior algorithms on tensor regression and tensor SVD is given in Table \ref{tab: comparison table}. We can see for fixed $r$, RGN achieves the best estimation error and signal-to-noise ratio requirement, i.e., sample complexity in tensor regression and least singular value in tensor SVD, compared to the state of the art while maintaining a relatively low computational cost. Moreover, RGN is the only algorithm with guaranteed quadratic convergence in both applications. Finally, we conduct numerical studies to support our theoretical findings in Section \ref{sec:numerics}. The simulation studies show RGN offers much faster convergence compared to the existing approaches in the literature. 

\begin{table}[ht]
	\centering
	\begin{tabular}{c |c | c | c | c }
	\hline
	& \multicolumn{4}{|c}{Tensor Regression} \\
	\hline
	\multirow{2}{4em}{Algorithm} & required & convergence & estimation  & per-iteration   \\
	 & sample size & rate &  error & cost  \\
	\hline
	RGN &  \multirow{2}{4em}{$p^{d/2} r^{3/2}$} & \multirow{2}{4em}{quadratic} & \multirow{2}{4em}{$\sigma\sqrt{\frac{pr}{n}}$}  & \multirow{2}{4em}{$np^d r$} \\
	(this work) & & & &\\
	\hline
	GD  &  \multirow{2}{4em}{$p^{d/2} r^{3/2}$} & \multirow{2}{4em}{linear} & \multirow{2}{4em}{$\sigma\sqrt{\frac{pr}{n}}$}   & \multirow{2}{4em}{$np^d$} \\
	\citep{han2020optimal} & & & &\\
	\hline
	Nonconvex-PGD &  \multirow{2}{4em}{$p^{d-1}r$} & \multirow{2}{4em}{linear} & \multirow{2}{4em}{$\sigma\sqrt{\frac{p^{d-1}}{n}}$}  & \multirow{2}{4em}{$np^d$} \\
	\citep{chen2016non} & & & & \\
	\hline
	Alter Mini &  \multirow{2}{4em}{N.A.} & \multirow{2}{4em}{linear} & \multirow{2}{4em}{N.A.} & \multirow{2}{4em}{$np^d r$}  \\
	\citep{zhou2013tensor} & & & &\\
	 	 \hline
	 & \multicolumn{4}{|c}{Tensor SVD} \\
	 	\hline
	 \multirow{2}{4em}{Algorithm}	 & least & convergence & estimation & per-iteration\\
	  & singular value & rate & error & cost\\

	 \hline
	RGN  & \multirow{2}{4em}{$p^{d/4} r^{1/4} \sigma$} & \multirow{2}{4em}{quadratic} & \multirow{2}{4em}{$\sigma \sqrt{pr}$} & \multirow{2}{4em}{$p^d r$}\\
	(this work) & & & &\\
	\hline
	GD & \multirow{2}{4em}{$p^{d/4} r^{1/4}\sigma$} & \multirow{2}{4em}{linear} & \multirow{2}{4em}{$\sigma \sqrt{pr}$} & \multirow{2}{4em}{$p^d r$} \\
	\citep{han2020optimal} & & & &\\
	\hline
	Alter Mini & \multirow{2}{4em}{$p^{d/4} \sigma$} & \multirow{2}{4em}{linear} & \multirow{2}{4em}{$\sigma \sqrt{pr}$} & \multirow{2}{4em}{$p^d r$} \\
	\citep{zhang2018tensor} & & & &\\
	\hline
	\end{tabular}
	\caption{ Comparison of RGN with gradient descent (GD), nonconvex projected gradient descent (Nonconvex-PGD) and alternating minimization (Alter Mini) in the literature in aspects of signal-to-noise ratio requirement (i.e., overall sample complexity in tensor regression and least singular value in tensor SVD), convergence rate, estimation error, and per-iteration computational cost. Here the convergence rate is global for nonconvex-PGD and is local for other algorithms. We assume the Tucker rank satisfies $r_1=\cdots=r_d=r$, the tensor dimension satisfies $ p_1=\cdots=p_d$, $ r,d\ll n,p$, the tensor parameter of interest is well-conditioned and $\sigma$ is the standard deviation of the Gaussian noise in tensor regression and tensor SVD.
	}\label{tab: comparison table}
\end{table}

\subsection{Related Literature} \label{sec: literature review}

Our work is related to a broad range of literature from a number of communities. Here we make an attempt to discuss existing results without claiming that the survey is exhaustive.

First, the low-rank tensor estimation has attracted much recent attention from machine learning and statistics communities. Various methods were proposed, including the convex relaxation \citep{mu2014square,raskutti2015convex,tomioka2011statistical}, projected gradient descent~\citep{rauhut2017low,chen2016non,ahmed2020tensor,yu2016learning}, gradient descent on the factorized model \citep{han2020optimal,cai2019nonconvex,hao2018sparse}, alternating minimization \citep{zhou2013tensor,jain2014provable,liu2020tensor,xia2017statistically}, and importance sketching \citep{zhang2020islet}. A scaled GD was proposed in the concurrent work \cite{tong2022scaling}. Moreover, when the target tensor has order two, our problem reduces to the widely studied low-rank matrix recovery/estimation \citep{recht2010guaranteed,li2019rapid,ma2019implicit,sun2015guaranteed,tu2016low,wang2017unified,zhao2015nonconvex,zheng2015convergent,charisopoulos2021low,luo2020recursive,bauch2021rank}. The readers are referred to a recent survey in \cite{chi2019nonconvex}. 

Second, Riemannian manifold optimization methods have been powerful in solving optimization problems with geometric constraints \citep{absil2009optimization,boumal2020introduction}. Many progresses in this topic were made for the low-rank matrix estimation \citep{keshavan2009matrix,boumal2011rtrmc,boumal2015low,wei2016guarantees,meyer2011linear,mishra2014fixed,vandereycken2013low,huang2018blind,cherian2016riemannian}. In particular, \cite{luo2020recursive} proposed a recursive importance sketching algorithm for solving rank-constrained least-squares problem and showed that it is closely related to the Riemannian Gauss-Newton method. See also the recent survey on this line of work at \cite{cai2018exploiting,uschmajew2020geometric}. Moreover, the Riemannian manifold optimization method has been applied for various problems on low-rank tensor estimation, such as tensor regression \citep{kressner2016preconditioned}, tensor completion \citep{rauhut2015tensor,kasai2016low,dong2021new,kressner2014low,heidel2018riemannian,xia2017polynomial,steinlechner2016riemannian,da2015optimization}, and robust tensor PCA \citep{cai2021generalized}. These papers mostly focus on the first-order Riemannian optimization methods, possibly due to the hardness of deriving the exact expressions of the Riemannian Hessian. A few exceptions also appear: \cite{heidel2018riemannian,kasai2016low} and \cite{psenka2020second} developed Riemannian trust-region method for tensor completion under Tucker and tensor-train formats, respectively; a Riemannian Gauss-Newton algorithm was also considered in \cite{heidel2018riemannian}; \cite{kressner2016preconditioned} proposed approximate Riemannian Newton methods for tensor regression in tensor-train and Tucker formats under the setting that the linear map has additive and Kronecker-product-type structures. Departing from these results that focus on the geometric objects and numerical implementations, in this paper we not only develop an efficient implementation of RGN under the Tucker format but also prove the quadratic convergence of the iterates and the optimal estimation error rate for the estimation of $\bcX^*$. 

Finally, the low-rank tensor estimation model \eqref{eq:model} is related to various problems in different contexts. In the tensor-based scientific computing community, large-scale linear systems where the solution admits a low-rank tensor structure commonly arise after discretizing high-dimensional partial differential equations (PDEs) \citep{lynch1964tensor,hofreither2018black}, which exactly become the central problem \eqref{eq:model} in this paper. In the literature, various methods have been proposed there to solve \eqref{eq:model}. For example, \cite{bousse2018linear} developed the algebraic method and Gauss-Newton method to solve the linear system with a CP low-rank tensor solution. \cite{georgieva2019greedy} and \cite{kressner2016preconditioned} respectively introduced a greedy approach and an approximate Riemannian Newton method to approximate the linear system by a low Tucker rank tensor. The readers are also referred to \cite{grasedyck2013literature} for a recent survey. There are some key differences from this line of work to ours: first, their goal is often to find a low-rank tensor that approximately solves a linear system with small {\it approximation error}, while we aim to develop an estimator with small {\it estimation error}; second, the design matrix in linear systems from discretized PDEs often has Kronecker-product-type structure, we do not assume such structure in this paper. On the other hand, the structures of the design assumed here are application dependent, e.g., sub-Gaussian design in tensor regression and ``one-hot'' design in tensor completion as we mentioned in the introduction; finally, their work mainly focuses on computational aspects of the proposed methods \citep{grasedyck2013literature}, while this paper develops Riemannian Gauss-Newton for solving the low-rank tensor estimation problem and gives theoretical guarantees for the quadratic convergence of the algorithm and for the optimal estimation error bound of the final estimator.

\subsection{Organization of the Paper} \label{sec: organization}
After a brief introduction of notation and preliminaries in Section \ref{sec:notations}, we introduce our main algorithm RGN and its geometric ingredients in Section \ref{sec: Algorithm}. The theoretical results of RGN and its applications in tensor regression and tensor SVD are discussed in Sections \ref{sec:theory} and \ref{sec: statistics applications}, respectively. The computational complexity of RGN and numerical studies are presented in Sections \ref{sec: computation complexity and implementation} and \ref{sec:numerics}, respectively. Conclusion and future work are given in Section \ref{sec: conclusion}. Additional algorithms and all technical proofs are presented in Appendices \ref{sec: HOSVD, STHOSVD}-\ref{sec: additional lemmas}.

\subsection{Notation and Preliminaries}\label{sec:notations}
The following notation will be used throughout this article. Lowercase letters (e.g., $a$), lowercase boldface letters (e.g., $\u$), uppercase boldface letters (e.g., $\U$), and boldface calligraphic letters (e.g., $\bcA$) are used to denote scalars, vectors, matrices, and order-3-or-higher tensors, respectively. For simplicity, we denote $\bcA_j$ as the tensor indexed by $j$ in a sequence of tensors $\{\bcA_j\}$. We use bracket subscripts to denote sub-vectors, sub-matrices, and sub-tensors. For any vector $\v$, define its $\ell_2$ norm as $\|\v\|_2 = \left(\sum_i |v_i|^2\right)^{1/2}$. For any matrix $\D \in \mathbb{R}^{p_1\times p_2}$, let $\sigma_k(\D)$ be the $k$th largest singular value of $\D$. We also denote $\SVD_r(\D) = [\u_1 ~ \cdots \u_r]$ and QR($\D$) as the subspace composed of the leading $r$ left singular vectors and the Q part of the QR orthogonalization of $\D$, respectively. 
$\I_r$ represents the $r$-by-$r$ identity matrix. Let $\mathbb{O}_{p, r} = \{\U: \U^\top \U=\I_r\}$ be the set of all $p$-by-$r$ matrices with orthonormal columns. For any $\U\in \mathbb{O}_{p, r}$, $P_{\U} = \U\U^\top$ represents the projection matrix onto the column space of $\U$; we use $\U_\perp\in \mathbb{O}_{p, p-r}$ to represent the orthonormal complement of $\U$. 

The matricization $\mathcal{M}_k(\cdot)$ is the operation that unfolds the order-$d$ tensor $\bcA\in\mathbb{R}^{p_1\times \cdots \times p_d}$ along mode $k$ into the matrix $\mathcal{M}_k(\bcA)\in \mathbb{R}^{p_k\times p_{-k}}$ where $p_{-k} = \prod_{j\neq k}p_j$. Specifically, the mode-$k$ matricization of $\bcA$ is formally defined as 
\begin{equation} \label{eq: tensor-matricization}
	\mathcal{M}_k(\bcA) \in \mathbb{R}^{p_k\times p_{-k}}, \quad \left(\mathcal{M}_k(\bcA)\right)_{\left[i_k, j\right]} =\bcA_{[i_1,\ldots, i_d]}, \quad j = 1 + \sum_{\substack{l=1\\l\neq k}}^d\left\{(i_l-1)\prod_{\substack{m=1\\m\neq k}}^{l-1}p_m\right\}
\end{equation}
for any $1\leq i_l \leq p_l, l=1,\ldots, d$. We also use notation $\cT_k(\cdot)$ to denote the mode-$k$ tensorization or reverse operator of $\cM_k(\cdot)$. Throughout the paper, $\cT_k$, as a reversed operation of $\cM_k(\cdot)$, maps a $\bbR^{p_k \times p_{-k}}$ matrix back to a $\bbR^{p_1 \times \cdots \times p_d}$ tensor. 
The Hilbert-Schmidt norm of $\bcA$ is defined as
	$\|\bcA\|_{\tHS} = \left( \langle \bcA, \bcA \rangle \right)^{1/2}.$ 
The Tucker rank of a tensor $\bcA$ is denoted by $\tuckerrank(\bcA)$ and defined as a $d$-tuple $\br := (r_1, \ldots, r_d)$, where $r_k = \text{rank}(\mathcal{M}_k(\bcA))$. For any Tucker rank-$(r_1,\ldots,r_d)$ tensor $\bcA$, it has Tucker decomposition \citep{tucker1966some}: 
\begin{equation}\label{eq:tucker-decomposition}
 	\bcA = \llbracket \bcS; \U_1, \ldots, \U_d\rrbracket := \bcS\times_1 \U_1 \times \cdots \times_d \U_d,
\end{equation}
where $\bcS \in \bbR^{r_1 \times \cdots \times r_d}$ is the core tensor and $\U_k = \SVD_{r_k}(\cM_k(\bcA))$ is the mode-$k$ singular vectors. Here, the mode-$k$ product of $\bcA \in \mathbb{R}^{p_1 \times \cdots \times p_d}$ with a matrix $\B\in \mathbb{R}^{r_k\times p_k}$ is denoted by $\bcA \times_k \B$ and is of size $p_1 \times \cdots \times p_{k-1}\times r_k \times p_{k+1}\times \cdots \times p_d$, and its formal definition is given below 
\begin{equation}\label{eq: tensor-matrix product}
	(\bcA \times_k \B)_{[i_1, \ldots, i_{k-1}, j, i_{k+1}, \ldots, i_d]} = \sum_{i_k=1}^{p_k} \bcA_{[i_1, i_2, \ldots, i_d]} \B_{[j, i_k]}.
\end{equation}
It is convenient to introduce the following abbreviations to denote the tensor-matrix product along multiple modes: $\bcA \times_{k=1}^d \U_k := \bcA \times_{1} \U_1 \times \cdots \times_{d} \U_d$; $\bcA \times_{l\neq k}\U_l := \bcA \times_{1} \U_{1} \times \cdots \times_{k-1} \U_{k-1} \times_{k+1} \U_{k+1} \times \cdots \times_{d} \U_{d}$.
	The following property about tensor matricization will be used \cite[Section 4]{kolda2001orthogonal}:
\begin{equation}\label{eq: matricization relationship}
	\mathcal{M}_k \left( \bcS\times_1 \U_1 \times \cdots \times_d \U_d \right) = \U_k \mathcal{M}_k(\bcS) (\U_d^\top \otimes \cdots \otimes \U_{k+1}^\top \otimes \U_{k-1}^\top \otimes \cdots \otimes \U_{1}^\top),
\end{equation}
	where ``$\otimes$'' is the matrix Kronecker product. For any tensor $\bcZ \in \bbR^{p_1 \times \cdots \times p_d}$, we define $\bcZ_{\max(\br)} := \bcZ \times_{k=1}^d P_{\widehat{\U}_k}$ as the best Tucker rank $\br$ approximation of $\bcZ$ in terms of Hilbert-Schmidt norm, where $(\widehat{\U}_1, \ldots, \widehat{\U}_d) = \argmax_{\U_k \in \bbO_{p_k, r_k}, k=1,\ldots,d} \|\bcZ \times_{k=1}^d P_{\U_k} \|_{\tHS}.$
Finally, for any linear operator ${\cal L}$, we denote $\cL^*$ as its adjoint operator.

\section{Algorithm}\label{sec: Algorithm}

We introduce the geometry of low Tucker rank tensor Riemannian manifolds in Section \ref{sec: tensor manifold geometry} and present the procedure of RGN in Section \ref{sec: RGN}. 

\subsection{The Geometry for Low Tucker Rank Tensor Manifolds}\label{sec: tensor manifold geometry}

Denote the collection of $(p_1,\dots, p_d)$-dimensional tensors of Tucker rank $\br$ by $\bbM_{\br}=\{\bcX \in \bbR^{p_1 \times \cdots \times p_d}, \tuckerrank (\bcX) = \br \}$. Then $\bbM_{\br}$ forms a smooth submanifold embedded in $\bbR^{p_1 \times \cdots \times p_d}$ with dimension $\prod_{j=1}^d r_j + \sum_{j=1}^d r_j(p_j - r_j)$ \citep{uschmajew2013geometry,kressner2014low}. Throughout the paper, we use the natural Euclidean inner product as the Riemannian metric. Suppose $\bcX \in \bbM_{\br}$ has Tucker decomposition $\llbracket \bcS; \U_1,\ldots, \U_d \rrbracket$;  \cite{koch2010dynamical} showed that the tangent space of $\bbM_{\br}$ at $\bcX$, $T_{\bcX}\bbM_{\br}$, can be represented as
\begin{equation}\label{eq: old-def-tangent space}
	T_{\bcX} \bbM_{\br} = \left\{\bcB \times_{k=1}^d \U_k + \sum_{k=1}^d \bcS \times_k \widebar{\D}_k \times_{j \neq k} \U_j : \begin{array}{l}
		\bcB \in \bbR^{r_1 \times \cdots \times r_d}, \widebar{\D}_k \in \bbR^{p_k\times r_k},\\
		\widebar{\D}_k^\top \U_k =\0, k=1,\ldots,d
	\end{array}\right\}.
\end{equation}
In the representation above, $\widebar{\D}_k$s are not free parameters due to the constraints $\widebar{\D}_k^\top \U_k =0$, $k=1,\ldots,d$. In the following Lemma \ref{lm: mini para of tang space}, we introduce another representation of $T_{\bcX} \bbM_{\br}$ with a minimal parameterization, which matches the degree of freedom of the tangent space ($\prod_{j=1}^d r_j + \sum_{j=1}^d r_j(p_j - r_j)$). For $\bcX = \llbracket \bcS; \U_1,\ldots, \U_d \rrbracket$, we let $\V_k=\QR(\cM_k(\bcS)^\top)$, which corresponds to the row space of $\cM_k(\bcS)$, and define
\begin{equation} \label{def: Wk}
	\W_k := (\U_d \otimes \cdots \otimes \U_{k+1} \otimes \U_{k-1} \otimes \U_1) \V_k \in \bbO_{p_{-k}, r_{k}}, \quad k = 1,\ldots, d,
\end{equation}
where $p_{-k} = \prod_{j\neq k} p_j$. By \eqref{eq: matricization relationship}, $\U_k,\W_k$ correspond to the subspaces of the column and row spans of $\cM_k(\bcX)$, respectively.
\begin{Lemma}\label{lm: mini para of tang space}
	The tangent space of $\bbM_{\br}$ at $\bcX = \llbracket \bcS; \U_1,\ldots, \U_d \rrbracket $ in \eqref{eq: old-def-tangent space} can be written as
\begin{equation*}
    T_{\bcX} \bbM_{\br} = \left\{  \bcB \times_{k=1}^d \U_k + \sum_{k=1}^d \cT_k(\U_{k\perp}\D_k \W_k^{\top}): \bcB \in \bbR^{r_1 \times \cdots \times r_d}, \D_k \in \bbR^{(p_k - r_k)\times r_k}, k=1,\ldots,d\right\},
\end{equation*} 
where $\cT_k(\cdot)$ is the mode-$k$ tensorization operator and $\W_k$ is given in \eqref{def: Wk}.
\end{Lemma} 

We can also show that any tensor in $T_{\bcX}\bbM_{\br}$ is at most Tucker rank $2\br$. This fact will facilitate the efficient computation of RGN to be discussed in Section \ref{sec: computation complexity and implementation}. 
\begin{Lemma}\label{lm: tangent vector rank 2r property}
Any tensor $\bcZ \in T_{\bcX}\bbM_{\br}$ is at most Tucker rank $2\br$.
\end{Lemma}

Lemma 3.1 of \cite{koch2010dynamical} and the tangent space representation in Lemma \ref{lm: mini para of tang space} yield the following projection operator $P_{T_{\bcX}}$ that projects any tensor $\bcZ$ onto the tangent space of $\bbM_\br$ at $\bcX$:
\begin{equation} \label{eq: tangent space projector}
	P_{T_{\bcX}} (\bcZ) := \cL \cL^*(\bcZ) = \bcZ \times_{k=1}^d P_{\U_k} + \sum_{k=1}^d \cT_k(P_{\U_{k\perp}} \cM_k(\bcZ) P_{\W_k}  ), \quad \forall \bcZ \in \bbR^{p_1 \times \cdots \times p_d},
\end{equation}
where $\cL^*$ and $\cL$ are respectively the contraction map and extension map defined as follows: 
\begin{equation}\label{eq: Lt}
\begin{split}
	&\cL: \bbR^{r_1 \times \cdots \times r_d } \times  \prod_{k=1}^d \bbR^{(p_k - r_k) \times r_k}  \to T_{\bcX} \bbM_{\br}, ~~ (\bcB, \{\D_k \}_{k=1}^d) \mapsto \bcB \times_{k=1}^d \U_k + \sum_{k=1}^d \cT_k(\U_{k\perp} \D_k \W_k^{\top}),\\
	&\cL^*: \bbR^{p_1 \times \cdots \times p_d}  \to \bbR^{r_1 \times \cdots \times r_d}\times  \prod_{k=1}^d \bbR^{(p_k - r_k) \times r_k}, ~~ \bcZ \mapsto (\bcZ \times_{k=1}^d \U_k^{\top}, \{\U_{k\perp}^{\top} \cM_k(\bcZ) \W_k\}_{k=1}^d).
\end{split}
\end{equation}
In particular, $\cL^*$ is the adjoint operator of $\cL$. We will see in Section \ref{sec: RGN} that the representation in \eqref{eq: tangent space projector} helps the efficient implementation of RGN. 

\subsection{Riemannian Optimization and Riemannian Gauss-Newton}\label{sec: RGN}

In this subsection, we first give a preliminary for Riemannian optimization and then introduce the procedure of RGN for low-rank tensor estimation.

\vskip.2cm

{\noindent\bf Overall three-step procedure of Riemannian optimization.} Riemannian optimization concerns optimizing a real-valued function $f$ defined on a Riemannian manifold $\bbM$, for which the readers are referred to \cite{absil2009optimization} and \cite{boumal2020introduction} for an introduction. Due to the common non-linearity, the continuous optimization on the Riemannian manifold often requires calculations on the tangent space. A typical procedure of a Riemannian optimization method contains three steps per iteration: Step 1. find the tangent space; Step 2. update the point on the tangent space; Step 3. map the point from the tangent space back to the manifold.

\vskip.2cm

{\noindent\bf Low-rank tensor Riemannian manifold (Step 1).} We have already discussed the tangent space of low Tucker rank tensor manifolds in Section \ref{sec: tensor manifold geometry}, i.e., Step 1 above.

\vskip.2cm

{\noindent\bf Update on tangent space (Step 2).} Next, we describe the procedure of RGN in the tangent space. 
We begin by introducing a few more preliminaries for Riemannian manifold optimization. The Riemannian gradient of a smooth function $f:\bbM_{\br} \to \bbR$ at $\bcX\in \bbM_{\br}$ is defined as the unique tangent vector ${\rm grad}\, f(\bcX) \in T_\bcX \bbM_{\br}$ such that $\langle {\rm grad}\, f(\bcX), \bcZ \rangle = {\rm D}\, f(\bcX)[\bcZ], \forall\, \bcZ \in T_\bcX \bbM_{\br},$ where ${\rm D}f(\bcX)[\bcZ]$ denotes the directional derivative of $f$ at point $\bcX$ along direction $\bcZ$. Specifically for the embedded submanifold $\bbM_{\br}$, we have:
\begin{Lemma}\label{lm:gradient}
For $f(\bcX)$ in \eqref{eq:minimization}, ${\rm grad}\, f(\bcX) = P_{T_\bcX}(\scA^*(\scA(\bcX) - \y)),$
where $P_{T_\bcX}(\cdot)$ is the projection operator onto the tangent space of $\bbM_{\br}$ at $\bcX$ defined in \eqref{eq: tangent space projector}.
\end{Lemma}

A common way to derive RGN update in the literature is to first write down the Riemannian Newton equation, then replace the Riemannian Hessian by its Gauss-Newton approximation \citep[Chapter 8.4.1]{absil2009optimization}, and finally solve the modified Riemannian Newton equation, i.e., the Riemannian Gauss-Newton equation. In our low-rank tensor estimation problem with the objective function \eqref{eq:minimization}, suppose the current iterate is $\bcX^t$, the RGN update $\eta^{RGN} \in T_{\bcX^t} \bbM_{\br}$ should solve the following RGN equation \citep[Chapter 8.4]{absil2009optimization},
\begin{equation} \label{eq: Riemannian Gauss-newton equation}
	-\grad f(\bcX^t)= P_{T_{\bcX^t}} \left( \scA^* (\scA (\eta^{RGN})) \right).
\end{equation}
However, it is not clear how to solve this equation directly in practice.

Inspired by the classical Gauss-Newton (GN) algorithm, we instead introduce another scheme to derive RGN. Recall in solving the nonlinear least squares problem in the Euclidean space $\min_x\frac{1}{2}\|h(x)\|^2_2$, the classic Gauss-Newton can be viewed as a modified Newton method, and can also be derived by replacing the non-linear function $h(x)$ by its local linear approximation at the current iterate $x_k$ \cite[Chapter 10.3]{nocedal2006numerical}. These two ways of interpretation are equivalent. A similar local linearization idea can be extended to the manifold setting except that the linearization needs to be taken in the tangent space in each iterate. 
Specifically, consider the objective function $f(\bcX)$ in \eqref{eq:minimization}, the linearization of $\y - \scA(\bcX)$ at $\bcX^t$ in $T_{\bcX^t} \bbM_{\br}$ is $\y -\scA(\bcX^t) -\scA P_{T_{\bcX^t}}(\bcX - \bcX^t)$, which can be simplified to $\y - \scA P_{T_{\bcX^t}}(\bcX)$. After we further constraint the update direction to $T_{\bcX^t} \bbM_{\br}$, we have
\begin{equation} \label{eq: RGN-2}
\begin{split}
	\bcZ^{t+1} =  \argmin_{\bcZ \in T_{\bcX^t} \bbM_{\br} } \frac{1}{2}\|\y - \scA P_{T_{\bcX^t}}(\bcZ)\|_2^2.
\end{split}
\end{equation}
By mapping $\bcZ^{t+1}$ back to the manifold, we get the new iterate $\bcX^{t+1}$.

Next, we show the proposed update derived in \eqref{eq: RGN-2} actually matches the standard RGN update \eqref{eq: Riemannian Gauss-newton equation}.
\begin{Proposition}\label{prop: Riemannian Gauss-Newton of RGN}
Let $\bcZ^{t+1}$ be the update computed in \eqref{eq: RGN-2}. Then, $\bcZ^{t+1} - \bcX^t$ is the Riemannian Gauss-Newton update, i.e., it solves the Riemannian Gauss-Newton equation \eqref{eq: Riemannian Gauss-newton equation}. 
\end{Proposition}
Proposition \ref{prop: Riemannian Gauss-Newton of RGN} shows that \eqref{eq: RGN-2} yields the RGN update, which directly provides a simple implementation of RGN. To see this, recall $P_{T_{\bcX^t}} = \cL_t \cL_t^*$, where $\cL_t$ and $\cL_t^*$ are defined in the similar way as in \eqref{eq: Lt} except evaluated on $\bcX^t = \llbracket \bcS^t; \U_1^t,\ldots, \U^t_d \rrbracket$; then the objective function in \eqref{eq: RGN-2} can be rewritten as follows,
\begin{equation}\label{eq: least-square decomposition}
\begin{split}
	& \frac{1}{2}\|\y - \scA P_{T_{\bcX^t}}(\bcZ)\|_2^2 = \frac{1}{2} \sum_{i=1}^n ( \y_i - \langle \bcA_i, \cL_t \cL_t^* \bcZ  \rangle )^2 = \frac{1}{2} \sum_{i=1}^n ( \y_i - \langle \cL_t^* \bcA_i, \cL_t^* \bcZ  \rangle )^2 \\
	& = \frac{1}{2} \sum_{i=1}^n \left( \y_i - \langle \bcA_i \times_{k=1}^d \U_k^{t\top}, \bcB \rangle - \sum_{k=1}^d \langle  \U_{k\perp}^{t\top}\mathcal{M}_k\left(\bcA_{i}\right) \W_k^t, \D_k \rangle  \right)^2,
\end{split}
\end{equation} 
where $(\bcB, \{ \D_k \}_{k=1}^d ):= \cL_t^* \bcZ$. Based on the calculation in \eqref{eq: least-square decomposition}, we define the following covariates maps $\sA_{\bcB}: \bbR^{r_1 \times \cdots \times r_d} \to \bbR^n, \sA_{\D_k}: \bbR^{(p_k-r_k)r_k} \to \bbR^n, k= 1,\cdots, d $, where for $1 \leq i \leq n$,
\begin{equation*}
	\begin{split}
		&(\sA_\bcB)_{i} = 
		\bcA_i \times_{k=1}^d \U_k^{t\top}, \quad (\sA_{\D_k})_{i} = \U_{k\perp}^{t\top}\mathcal{M}_k\left(\bcA_{i}\right) \W_k^t.
	\end{split}
\end{equation*}
Here, $(\scA_\bcB)_i$ satisfies $[\scA_\bcB(\cdot)]_i = \langle \cdot, (\scA_\bcB)_i\rangle$ and similarly for $(\scA_{\D_k})_i$. Then, by \eqref{eq: least-square decomposition} and the fact that $\bcZ \in T_{\bcX^t} \bbM_{\br}$, \eqref{eq: RGN-2} can be equivalently solved by
\begin{equation*}
\begin{split}
	\bcZ^{t+1} &= \cL_t (\B^{t+1},\D_1^{t+1},\cdots,\D_d^{t+1}),
\end{split}
\end{equation*}
where 
\begin{equation}\label{eq: modified least-square}
	(\bcB^{t+1}, \D_1^{t+1}, \cdots, \D_d^{t+1}) = \argmin_{\substack{\bcB\in \bbR^{r_1 \times \cdots \times r_d}, \\ \D_k\in \bbR^{(p_k -r_k) \times r_k}, k=1,\ldots,d}} \left\|\y - \sA_\bcB(\bcB) - \sum_{k=1}^d\sA_{\D_k}(\D_k) \right\|_2^2.
\end{equation}
Note that \eqref{eq: modified least-square} is an unconstrained least squares with the number of parameters equal to $\sum_{k=1}^d (p_k-r_k)r_k + \prod_{k=1}^d r_k$.

\vskip.2cm

{\noindent\bf Retraction (Step 3).} 
Finally, we discuss how to map the point from the tangent space back to the manifold, i.e., Step 3 above. An ideal method is via the {\it exponential map}, which moves a point on the manifold along the geodesic. However, computing the exponential map is prohibitively expensive in most situations, and a more practical choice is the so-called {\it retraction}. Retraction is in general a first-order approximation of the exponential map. In tensor manifold $\bbM_{\br}$, the retraction map, denoted by $R$, should be a smooth map from $T\bbM_{\br}$ to $\bbM_{\br}$ that satisfies i) $R(\bcX, 0) = \bcX$ and ii) $\frac{d}{d t} R(\bcX, t \eta) \vert_{t = 0} = \eta$ for all $\bcX \in \bbM_{\br}$ and $\eta \in T_\bcX \bbM_{\br}$ \citep[Chapter 4]{absil2009optimization}. Here, $T \bbM_{\br} = \{(\bcX, T_\bcX \bbM_{\br}) : \bcX \in \bbM_{\br}\}$ is the tangent bundle of $\bbM_{\br}$.

In the low Tucker rank tensor manifolds, Proposition 2.3 of \cite{kressner2014low} showed that the truncated high-order singular value decomposition (T-HOSVD) \citep{de2000multilinear} is a retraction. We further show in the following Lemma \ref{lm: st-HOSVD as retraction} that the sequentially truncated HOSVD (ST-HOSVD)  \citep{vannieuwenhoven2012new}, a computationally more efficient procedure than T-HOSVD, also satisfies the retraction properties. The detailed procedures of T-HOSVD and ST-HOSVD are given in Appendix \ref{sec: HOSVD, STHOSVD}.

\begin{Lemma}[Retraction of Sequentially Truncated HOSVD] \label{lm: st-HOSVD as retraction}
	For ST-HOSVD defined in Appendix \ref{sec: HOSVD, STHOSVD}, the map 
	\begin{equation*}
		R: T\bbM_{\br} \to \bbM_{\br}, (\bcX, \eta) \to {\text {\rm ST-HOSVD}}(\bcX + \eta)
	\end{equation*}
	is a retraction on $\bbM_{\br}$ around $\bcX$.
\end{Lemma}
Although ST-HOSVD has been widely used in practice, the retraction property of ST-HOSVD we established in Lemma \ref{lm: st-HOSVD as retraction} is new.

\vskip.2cm

{\bf\noindent Summary of RGN.} We give the complete RGN algorithm for low-rank tensor estimation in Algorithm \ref{alg:RGN}. 

\begin{algorithm}[h]
	\caption{Riemannian Gauss-Newton for Low-rank Tensor Estimation}
	\noindent {\bf Input}: $\y\in \mathbb{R}^n, \bcA_1,\ldots, \bcA_n \in \mathbb{R}^{p_1\times \cdots \times p_d}$, $t_{\max}$, Tucker rank $\br$, initialization $\bcX^0$ with Tucker decomposition $\llbracket \bcS^0; \U_1^0, \ldots, \U_d^0 \rrbracket$, and $\W_k^0$ defined as \eqref{def: Wk}.
	\begin{algorithmic}[1]
		\For{$t=0, 1, \ldots, t_{\max}-1$}
		\State Construct the covariates maps $\sA_{\bcB}: \bbR^{r_1 \times \cdots \times r_d} \to \bbR^n, \sA_{\D_k}: \bbR^{(p_k-r_k)r_k} \to \bbR^n, k= 1,\cdots, d $, where for $1 \leq i \leq n$
		\begin{equation}\label{eq: importance sketches}
		\begin{split}
		&(\sA_\bcB)_{i} = 
		\bcA_i \times_{k=1}^d \U_k^{t\top}, \quad (\sA_{\D_k})_{i} = \U_{k\perp}^{t\top}\mathcal{M}_k\left(\bcA_{i}\right) \W_k^t.
		\end{split}
		\end{equation}	
		\State Solve the unconstrained least squares problem
		\begin{equation} \label{eq: alg least square}
		(\bcB^{t+1}, \D_1^{t+1}, \cdots, \D_d^{t+1}) = \argmin_{\substack{\bcB\in \bbR^{r_1 \times \cdots \times r_d}, \\ \D_k\in \bbR^{(p_k -r_k) \times r_k}, k=1,\ldots,d}} \left\|\y - \sA_\bcB(\bcB) - \sum_{k=1}^d\sA_{\D_k}(\D_k) \right\|_2^2.
		\end{equation}
		\State Update
		\begin{equation}\label{eq: alg Xt update}
			\begin{split}
				\bcX^{t+1} = \llbracket \bcS^{t+1}; \U_1^{t+1}, \ldots, \U_d^{t+1} \rrbracket = \cH_{\br} \left(\bcB^{t+1} \times_{k=1}^d \U^t_k + \sum_{k=1}^d \cT_k(\U_{k\perp}^{t}\D_k^{t+1}\W_k^{t\top})\right)
			\end{split}
		\end{equation} 
		and $\W_k^{t+1}$ via \eqref{def: Wk}. Here $\cH_{\br} (\cdot)$ is the retraction map onto $\bbM_{\br}$ (two choices are ST-HOSVD and T-HOSVD).
		\EndFor
	\end{algorithmic}
	\noindent {\bf Output}: $\bcX^{t_{\max}}$.
	\label{alg:RGN}
\end{algorithm}

\begin{Remark}[Operator $\cH_{\br}$]\label{rem: quasi-projection} In \eqref{eq: alg Xt update}, $\cH_{\br}$ plays the role of retraction that maps the iterate from the tangent space of $\bbM_{\br}$ at $\bcX^t$ back onto the manifold. Since $\cH_{\br}$ directly operates on the updated tensor, to distinguish with the canonical notation $R(\cdot, \cdot)$ for retraction, we use a simplified notation $\cH_{\br}$ to represent this map here. As we mentioned before, T-HOSVD \citep{de2000multilinear} and ST-HOSVD \citep{vannieuwenhoven2012new} are two choices of retractions. 
\end{Remark}

\section{Theoretical Analysis}\label{sec:theory}

We analyze the convergence rate of RGN in this section. 

\subsection{Quasi-projection Property and Tensor Restricted Isometry Property }
We begin by introducing the quasi-projection property of T-HOSVD and ST-HOSVD and the assumption on the linear map $\scA$. Different from the low-rank matrix projection, which can be efficiently and exactly computed via truncated SVD, performing low-rank tensor projection exactly, even for $\br=1$, can be NP-hard in general \citep{hillar2013most}. We thus introduce the following quasi-projection property and the approximation constant $\delta(d)$. 

\begin{Definition}[Quasi-projection of $\cH_{\br}$ and Approximation Constant $\delta(d)$]\label{def: quasi-projection Hr}
Let $P_{\bbM_{\br}}(\cdot)$ be the projection map from $\bbR^{p_1 \times \cdots \times p_d}$ to the tensor space of Tucker rank at most $\br$, i.e., for any $\bcZ\in \mathbb{R}^{p_1\times\cdots\times p_d}$ and $\widehat{\bcZ}$ of Tucker rank at most $\br$, one always has $\|\bcZ - \widehat{\bcZ}\|_{\tHS}\geq\|\bcZ - P_{\bbM_{\br}}(\bcZ)\|_{\tHS}$. 

We say $\cH_{\br}$ satisfies the quasi-projection property with approximation constant $\delta(d)$ if $\|\bcZ - \cH_{\br}(\bcZ)\|_{\tHS} \leq \delta(d)\|\bcZ - P_{\bbM_{\br}}(\bcZ) \|_{\tHS}$ for any $\bcZ \in \bbR^{p_1 \times \cdots \times p_d}$. 
\end{Definition}
It is known that T-HOSVD and ST-HOSVD satisfy the \emph{quasi-projection property} (see Chapter 10 in \citep{hackbusch2012tensor}). 
\begin{Proposition}[Quasi-projection property of T-HOSVD and ST-HOSVD] \label{prop: quasi-projection}
    T-HOSVD and ST-HOSVD described in Appendix \ref{sec: HOSVD, STHOSVD} satisfy the quasi-projection property with approximation constant $\delta(d)=\sqrt{d}$.
\end{Proposition}

For technical convenience, we also assume $\scA$ satisfies the following Tensor Restricted Isometry Property (TRIP) \citep{rauhut2017low}. One major reason we need this assumption is to control the spectrum of the operator $\cL_t^* \scA^* \scA \cL_t$ presented in Lemma 6 in Appendix \ref{sec:proof} of the paper. The TRIP condition can be seen as a tensor generalization of the restricted isometry property (RIP). In the compressed-sensing and low-rank matrix recovery literature, the RIP condition has been widely used as one standard assumption \citep{candes2011tight,cai2013sharp}. 

\begin{Definition}[Tensor Restricted Isometry Property (TRIP)]\label{def: RIP}
Let $\scA: \bbR^{p_1 \times \cdots \times p_d} \to \bbR^n$ be a linear map. For a fixed $d$-tuple $\br = (r_1,\ldots,r_d)$ with $1 \leq r_k \leq p_k$ for $k=1,\ldots,d$, define the $\br$-tensor restricted isometry constant to be the smallest number $R_{\br}$ such that $(1-R_{\br}) \|\bcZ\|^2_{\tHS} \leq \|\scA(\bcZ)\|_2^2 \leq (1+R_{\br}) \|\bcZ\|_{\tHS}^2$ holds for all $\bcZ$ of Tucker rank at most $\br$. If $0\leq R_{\br} < 1$, we say $\scA$ satisfies $\br$-tensor restricted isometry property ($\br-$TRIP).
\end{Definition}
In \cite{rauhut2017low}, the authors showed that TRIP can be satisfied in a number of different scenarios. For example, if sensing tensors $\bcA_i$ are composed of i.i.d. sub-Gaussian entries, then with high probability, the TRIP condition can be satisfied with TRIP constant $R_{\br}$ as long as $n \geq C ( \sum_{k=1}^d p_k r_k + \prod_{k=1}^d r_k )\log d/R_{\br}^2$ for some constant $C > 0$. In addition, TRIP also holds for more structured measurement ensembles such as the random Fourier mapping \citep{rauhut2017low}.

\subsection{Main Convergence Results}
In this subsection, we establish the deterministic convergence theory for RGN. 
\begin{Theorem}[Convergence of RGN]\label{th: local contraction general setting}
Suppose $\cH_{\br}$ is either T-HOSVD or ST-HOSVD, $\scA$ satisfies the {$3\br$}-TRIP, and the initialization $\bcX^0$ satisfies $	\|\bcX^0 - \bcX^*\|_{\tHS} \leq \frac{\underline{\lambda} }{4d (\sqrt{d} + 1)(R_{3\br}/(1-R_{2\br}) + 1)}$,
where $\underline{\lambda}:= \min_{k=1,\ldots,d}\sigma_{r_k}(\cM_k(\bcX^*))$ is the minimum of least singular values at each matricization of $\bcX^*$. Then for all $t\ge 0$,
\begin{equation*}
    \|\bcX^{t+1} - \bcX^* \|_{\tHS} \leq d(\sqrt{d} + 1)\left(\frac{R_{3\br}}{1-R_{2\br}} + 1\right) \frac{\|\bcX^t - \bcX^*\|^2_{\tHS}}{\underline{\lambda}} + \frac{\sqrt{d} + 1}{1 - R_{2\br}} \|(\scA^*(\bvarepsilon ))_{\max(2\br)}\|_{\tHS}.
\end{equation*} 
Recall, $(\cdot)_{\max(\br)}$ denotes the best Tucker rank $\br$ approximation of the tensor ``$\cdot$".

In particular, if $\bvarepsilon = 0$, then $\{\bcX^t\}$ converges quadratically to $\bcX^*$ as
\begin{equation*}
	\|\bcX^{t+1} - \bcX^* \|_{\tHS}\leq d(\sqrt{d} + 1)\left(\frac{R_{3\br}}{1-R_{2\br}} + 1\right) \frac{\|\bcX^t - \bcX^*\|^2_{\tHS}}{\underline{\lambda}}, \quad \forall\, t \ge 0.
\end{equation*}
\end{Theorem}

	Theorem \ref{th: local contraction general setting} shows with some proper assumptions on $\scA$ and initialization, the iterates of RGN converge quadratically to the ball centered at $\bcX^*$ and of radius $O(\|(\scA^*(\bvarepsilon ))_{\max(2\br)}\|_{\tHS})$. Especially if $\bvarepsilon = 0$, i.e., the observations are noiseless, $\bcX^t$ converges quadratically to the exact $\bcX^*$. To the best of our knowledge, this is the first provable quadratic convergence guarantee for both low-rank tensor estimation and recovery. 
	
	We note that, in the noisy setting, RGN in general converges to a stationary point with nonzero function value, so the classical optimization theory can only yield local (super)linear convergence guarantee for RGN \citep{absil2009optimization,breiding2018convergence}. Our result complements the classic theory of RGN: \cite{absil2009optimization,breiding2018convergence} studied the limiting convergence rate of RGN to a stationary point, while we show RGN converges quadratically to $\bcX^*$, the true parameter of interest, up to some optimal statistical error (see the forthcoming Theorem \ref{th: pertur lower bound}). This also suggests that to achieve quadratic convergence performance in low-rank tensor estimation, the more sophisticated Riemannian Newton algorithm may be unnecessary as simple RGN already enjoys the quadratic convergence for estimating $\bcX^*$ with theoretical guarantees. 

\begin{Remark}[Initialization] \label{rem: initialization}
	The convergence theory in Theorem \ref{th: local contraction general setting} requires an initialization condition. Our condition says that $\|\bcX^0 - \bcX^*\|_{\tHS}$ needs to be on the order of $\underline{\lambda}$. In the matrix setting, i.e., $d = 2$, this condition matches the initialization condition in the literature for using two-stage nonconvex optimization methods to solve the low-rank matrix recovery problems \citep{charisopoulos2021low,ma2019implicit,sun2015guaranteed,tu2016low,wang2017unified,zhao2015nonconvex,zheng2015convergent}. To this point of view, our initialization condition is an extension of that. Moreover, in practice, the SVD-based methods often provide a sufficiently good initialization that meets the requirement in many statistical applications. We will further illustrate this point in Section \ref{sec: statistics applications}. The numerical studies in Section \ref{sec: numerical RGN property} show that RGN can still work well under random initialization. We leave future work to provide convergence guarantees of RGN under random initialization.
	
	 Moreover, there is also a factor $1/d^{3/2}$ in the initialization condition. In typical applications, such as brain MRI images or fMRI images, $d$ will be a moderately large value, say $3$ or $4$. If $d$ is large, then the $1/d^{3/2}$ factor will be important and the power of $d$ also comes for reasons. The first reason is due to the fact that performing exact low-rank tensor projection is computationally intractable and efficient procedures such as T-HOSVD and ST-HOSVD can only achieve quasi-projection property with approximation constant $\sqrt{d}$ as we discussed in Proposition \ref{prop: quasi-projection}. The second reason is from the result in Lemma \ref{lm: orthogonal projection} that it has a factor $d$ in the numerator of the bound. This can also be seen from the proof that the effective dimension of the orthogonal complement of $T_{\bcX^t}\bbM_{\br}$, which controls the error, scales linearly in $d$. These two factors together yield the $1/d^{3/2}$ factor.
 
\end{Remark}

\begin{Remark}[Convergence under Restricted Isometry Property] 
	In the literature, the RIP-type assumptions are widely used to establish linear convergence guarantees for various first-order algorithms in low-rank matrix/tensor recovery. A common strategy to establish such results is to first show the linear convergence of the empirical loss and then transfer to the convergence of the iterates \citep{jain2010guaranteed}. To our best knowledge, we are the first to use RIP to establish the second-order convergence of RGN by directly showing the contraction of the iterates. The key lemmas in our theoretical analysis are Lemma \ref{lm: bound for iter approx error}, which bounds the per-iteration least squares estimation error, and Lemma \ref{lm: orthogonal projection}, which bounds the projection of $\bcX^*$ on the orthogonal complement of $T_{\bcX^t} \bbM_{\br}$.
\end{Remark}

\begin{Corollary}[Two Phases of Convergence of RGN] \label{coro: two-phase convergence} 
Suppose the conditions in Theorem \ref{th: local contraction general setting} hold. Define $\Delta := \sqrt{\frac{ \underline{\lambda} \|(\scA^*(\bvarepsilon ))_{\max(2\br)}\|_{\tHS} }{  d(1+R_{3\br}-R_{2\br}) }}$. At iteration $t$,
\begin{itemize}
	\item[(Phase I)] If $\|\bcX^t - \bcX^*\|_{\tHS} \geq \Delta$, then $\|\bcX^{t+1} - \bcX^* \|_{\tHS} \leq  2d(\sqrt{d} + 1)(\frac{R_{3\br}}{1-R_{2\br}} + 1) \frac{\|\bcX^t - \bcX^*\|^2_{\tHS}}{\underline{\lambda}};$
	\item[(Phase II)] If $\|\bcX^t - \bcX^*\|_{\tHS} \leq \Delta$, then $ \|\bcX^{t+1} - \bcX^* \|_{\tHS} \leq \frac{2(\sqrt{d} + 1)}{1 - R_{2\br}} \|(\scA^*(\bvarepsilon ))_{\max(2\br)}\|_{\tHS}.$
\end{itemize}
	In summary, we have
	\begin{equation*}
		\|\bcX^t - \bcX^*\|_{\tHS} \leq 2^{-2^{t}} \|\bcX^0 - \bcX^*\|_{\tHS} + \frac{2(\sqrt{d} + 1)}{1 - R_{2\br}} \|(\scA^*(\bvarepsilon ))_{\max(2\br)}\|_{\tHS}, \quad \forall t \geq 0.
	\end{equation*}
	In addition, as long as 
	 \begin{equation}\label{eq: iteration complexity}
	 	t_{\max} \geq T_{\max}:= \left\lceil \log \left( 1 \vee \frac{1}{2} \log \left( \frac{d(1+R_{3\br}-R_{2\br}) \|\bcX^0 - \bcX^*\|_{\tHS}^2}{\underline{\lambda} \|(\scA^*(\bvarepsilon ))_{\max(2\br)}\|_{\tHS} }  \right)  \right) \right\rceil + 1,
	 \end{equation} 
	 we have $\|\bcX^{t_{\max}} - \bcX^* \|_{\tHS} \leq \frac{2(\sqrt{d} + 1)}{1 - R_{2\br}} \|(\scA^*(\bvarepsilon ))_{\max(2\br)}\|_{\tHS}.$
\end{Corollary}

Notice the initialization error required in Theorem \ref{th: local contraction general setting} is of order $\underline{\lambda}$, which is in general larger than $\|(\scA^*(\bvarepsilon ))_{\max(2\br)}\|_{\tHS}$, so the magnitude of $\Delta$ in Corollary \ref{coro: two-phase convergence} can be much bigger than $\|(\scA^*(\bvarepsilon ))_{\max(2\br)}\|_{\tHS}$. Corollary \ref{coro: two-phase convergence} shows that the convergence of RGN has two different phases: in Phase I that $\|\bcX^t - \bcX^*\|_{\tHS}$ is bigger than the threshold $\Delta$, $\bcX^t$ converges quadratically to $\bcX^*$; in Phase II that $\|\bcX^t - \bcX^*\|_{\tHS}$ is smaller than the threshold, with one extra step, the estimation error of $\bcX^{t+1}$ becomes at most $\frac{2(\sqrt{d} + 1)}{1 - R_{2\br}} \|(\scA^*(\bvarepsilon ))_{\max(2\br)}\|_{\tHS}$. 

\subsection{Optimality of RGN}
	
	Next, we further introduce a lower bound to show $\xi := \|(\scA^*(\bvarepsilon ))_{\max(2\br)}\|_{\tHS}$ is essential in the estimation error upper bounds of Theorem \ref{th: local contraction general setting} and Corollary \ref{coro: two-phase convergence}. 
	\begin{Theorem}[Minimax Lower Bound for Tensor Estimation]\label{th: pertur lower bound}
	Consider the following class of $(\widetilde{\scA}, \widetilde{\bcX}, \widetilde{\bvarepsilon})$:
	\begin{equation*}
	\begin{split}
	\mathcal{F}_{\br}(\xi) = \left\{ \left(\widetilde{\scA}, \widetilde{\bcX}, \widetilde{\bvarepsilon} \right): \begin{array}{l}
	\widetilde{\scA} \text{ satisfies } 3\br \text{-TRIP}, \widetilde{\bcX} \text{ is of Tucker rank at most } \br,\\ 
	\|(\widetilde{\scA}^*(\widetilde{\bvarepsilon} ))_{\max(2\br)}\|_{\tHS} \leq \xi
	\end{array}\right\}.
	\end{split}
	\end{equation*}
	Under the low-rank tensor estimation model \eqref{eq:model}, we have 
		$$\inf_{\widehat{\bcX}} \sup_{\left(\widetilde{\scA}, \widetilde{\bcX}, \widetilde{\bvarepsilon} \right) \in \mathcal{F}_{\br}(\xi)} \|\widehat{\bcX} - \widetilde{\bcX}\|_{\tHS} \geq 2^{-1/2} \xi.$$
	\end{Theorem}
	\begin{Remark}[(One-step) Optimality of RGN for Low-rank Tensor Estimation]\label{rem: optimality of RGN}
		Theorem \ref{th: pertur lower bound} and Corollary \ref{coro: two-phase convergence} together show that with a fixed tensor order $d$ and a proper initialization, RGN achieves rate-optimal estimation error in class $\mathcal{F}_{\br}(\xi)$ after at most double-logarithmic, i.e., $T_{\max}$ defined in \eqref{eq: iteration complexity}, number iterations. 
		
	    Corollary \ref{coro: two-phase convergence} also shows in Phase II convergence of RGN, with one extra step, the estimation error $\bcX^{t+1}$ becomes statistical rate-optimal. Such the one-step optimality shares the same spirit as the ``one-step MLE'' property for Newton algorithm in the literature on statistical inference (see \cite{bickel1975one} and \cite[Chapter 4.5]{shao2006mathematical}). To our best knowledge, this one-step optimality phenomenon is new for RGN.
	\end{Remark}

\section{Implications in Statistics and Machine Learning}\label{sec: statistics applications}

In this section, we study the performance of RGN in two specific problems in machine learning: tensor regression and tensor SVD. In addition, the algorithm is applicable to a broader range of settings discussed in the introduction. Throughout this section, we denote $\bar{p}:= \max_{k} p_k, \underline{p}:= \min_{k} p_k, \bar{r} = \max_k r_k $, $\underline{\lambda} := \min_{k} \sigma_{r_k}(\cM_k(\bcX^*))$, $\bar{\lambda} := \max_{k} \sigma_{1}(\cM_k(\bcX^*))$ and $\kappa := \bar{\lambda}/\underline{\lambda}$.

\subsection{Tensor Regression}\label{sec:tensor-regression}

Tensor Regression is a basic problem for supervised tensor learning, for which the readers are referred to Section \ref{sec: literature review} for a literature review. Specifically, we assume $\{\bcA_i\}_{i=1}^n$ are independent and have i.i.d. $N(0,1/n)$ entries; $\bvarepsilon_i \overset{i.i.d.}\sim N(0, \sigma^2/n)$ in model \eqref{eq:model}. 
Suppose the initialization is obtained by T-HOSVD:
\begin{equation} \label{eq: tensor-reg-initialization}
	\bcX^0 = \scA^*(\y)\times_{k=1}^d P_{\U_k^0},
\end{equation}
 where $\U_k^0 = \SVD_{r_k}(\cM_k(\scA^*(\y)))$. Then we have the following theoretical guarantee for the outcome of RGN for tensor regression. 

\begin{Theorem}[RGN for Tensor Regression] \label{th: tensor regression}
	Consider RGN for tensor regression. Suppose $\bar{r} \leq \underline{p}^{1/2}$, $\cH_{\br}$ is either T-HOSVD or ST-HOSVD.
	If $n \geq c(d) (\|\bcX^*\|_{\tHS}^2 + \sigma^2) \kappa^2 \sqrt{\bar{r}} \bar{p}^{d/2}/\underline{\lambda}^2 $, and $t_{\max} \geq C(d) \log\log\left( \frac{\underline{\lambda} \sqrt{n} }{\sigma\sqrt{\sum_{k=1}^d r_k p_k + \prod_{k=1}^d r_k}}\right)$, then
	\begin{equation} \label{ineq: tensor-regression-final-error}
		\|\bcX^{t_{\max}} - \bcX^*\|_{\tHS} \leq c(\sqrt{d} + 1)  \sigma\sqrt{\left(\sum_{k=1}^d r_k p_k + \prod_{k=1}^d r_k\right)/n}
	\end{equation}
	holds with probability at least $1 -  \underline{p}^{-C}$. Here $c,C$ are some universal positive constants, and $c(d), C(d)$ are some constants that depend on $d$ only.
\end{Theorem}

Suppose $d,\bar{r}$ are fixed. Note that $O(d\bar{p}\bar{r})$ samples is enough to guarantee the $\br$-TRIP and indeed this is the information-theoretic limit to make the problem solvable \citep[Theorem 5]{zhang2020islet}. However, we find to achieve the initialization assumption in Theorem \ref{th: local contraction general setting} via spectral method, a significantly larger sample complexity $O(c(d)\sqrt{\bar{r}} \bar{p}^{d/2} )$ is needed. Such a gap originates from the difficulty of computing the best Tucker rank $\br$ approximation of $\scA^*(\y)$ efficiently in initialization \citep{hillar2013most}. This difficulty is also a common reason that causes the so-called ``statistical and computational gaps" in various tensor problems \citep{richard2014statistical,zhang2018tensor,barak2016noisy,luo2020open,luo2020tensor,brennan2020reducibility,han2020exact}. See more discussions in Section \ref{sec: result-discussion}.

\subsection{Tensor SVD}\label{sec:tensor-svd}

Tensor SVD is a specific model covered by the prototypical model \eqref{eq:model}, which can be equivalently written as
$$\bcY = \bcX^* + \bcE,$$
where $\bcX^*$ has Tucker decomposition as \eqref{eq: decomposition of X} and $\bcE$ has i.i.d. $N(0,\sigma^2)$ entries. The goal is to estimate $\bcX^*$ based on $\bcY$. As illustrated by the following Lemma \ref{lm: RGN in ten-decom}, the RGN algorithm for tensor SVD can be significantly simplified from the original Algorithm \ref{alg:RGN}. 
\begin{Lemma}[Least Squares Solution of RGN in tensor SVD]\label{lm: RGN in ten-decom}
	Consider the tensor SVD model $\bcY = \bcX^* + \bcE$. Suppose at the $t$-th iteration of RGN, the current iterate is $\bcX^t$ with Tucker decomposition $\llbracket \bcS^{t}; \U_1^{t}, \ldots, \U_d^{t} \rrbracket$. Then the least squares in \eqref{eq: alg least square} can be solved by
	$$\bcB^{t+1} = \bcY \times_{k=1}^d \U_k^{t\top}, \quad \D_k^{t+1} = \U^{t\top}_{k\perp} \cM_k(\bcY) \W_k^t,\quad k = 1, \cdots, d.$$ 
	Here $\W_k^t$ is given in \eqref{def: Wk}.
\end{Lemma}
Consequently, we simplify RGN for tensor SVD to the following Algorithm \ref{alg:Iter ISLET-Ten Decom}.
	\begin{algorithm}[htbp]
	\caption{Riemannian Gauss-Newton for Tensor SVD}
	\noindent {\bf Input}: $\bcY \in \bbR^{p_{1} \times \cdots \times p_{d}}$, $t_{\max}$, a Tucker rank $\br$ initialization $\bcX^0$ with Tucker decomposition $\llbracket \bcS^0; \U_1^0, \ldots, \U_d^0 \rrbracket$.
	\begin{algorithmic}[1]
		\For{$t=0, 1, \ldots, t_{\max}-1$}
		\State Update
		\begin{equation*}
			\bcX^{t+1} = \llbracket \bcS^{t+1}; \U_1^{t+1}, \ldots, \U_d^{t+1} \rrbracket = \cH_{\br} \left(P_{T_{\bcX^t}}(\bcY) \right).	
		\end{equation*} Here $P_{T_{\bcX^t}}(\cdot)$ is the projection operator that project $\bcY$ onto the tangent space of $\bcX^t$ defined in \eqref{eq: tangent space projector}, $\cH_{\br} (\cdot)$ maps the input tensor to be a Tucker rank $\br$ tensor.
		\EndFor
	\end{algorithmic}
	\noindent {\bf Output}: $\bcX^{t_{\max}}$.
	\label{alg:Iter ISLET-Ten Decom}
\end{algorithm}

The following Theorem \ref{th: tensor SVD} gives the theoretical guarantee of RGN initialized with T-HOSVD for the tensor SVD.
\begin{Theorem}[RGN for Tensor SVD]\label{th: tensor SVD}
	Consider RGN for tensor SVD. Suppose $\bar{r}\leq \underline{p}^{1/2}$, $\cH_{\br}$ is either T-HOSVD or ST-HOSVD, and the algorithm is initialized by T-HOSVD, i.e., $\bcX^0 = \bcY \times_{k=1}^d P_{\U_k^0}$ where $\U_k^0 = \SVD_{r_k}(\cM_k(\bcY))$. If the least singular value $\underline{\lambda}\geq c(d)\kappa \bar{p}^{d/4}\bar{r}^{1/4}\sigma$, and $t_{\max} \geq C(d) \log \log \left\{ \underline{\lambda}\Big/\left(\sigma\sqrt{\sum_{k=1}^d r_k p_k + \prod_{k=1}^d r_k}\right)\right\}$,
	\begin{equation} \label{ineq: tensor-svd-final-error}
		\|\bcX^{t_{\max}} - \bcX^*\|_{\tHS} \leq c\cdot(\sqrt{d} + 1)  \sigma\sqrt{\sum_{k=1}^d r_k p_k + \prod_{k=1}^d r_k }
	\end{equation} 
	holds with probability at least $1 - \exp(-C \underline{p})$. 
\end{Theorem}

\begin{Remark}{\bf (RGN Versus Existing Algorithms in Tucker Tensor Decomposition)}
	We note that Algorithm \ref{alg:Iter ISLET-Ten Decom} can also be viewed as a Riemannian Gauss-Newton algorithm for Tucker tensor decomposition \citep{kolda2009tensor}. In the literature, a few other second-order methods have been proposed for Tucker tensor decomposition under Grassmann or quotient manifold structures, such as (quasi-)Newton-Grassmann method \citep{elden2009newton,savas2010quasi}, geometric Newton method \citep{ishteva2009differential}, and Riemannian trust region method \citep{ishteva2011best}. To the best of our knowledge, this is the first Riemannian Gauss-Newton algorithm for Tucker tensor decomposition developed using the embedded manifold structure on $\bbM_{\br}$. At the same time, we comment that different from the common goal in tensor decomposition which aims to find a good low-rank approximation of $\bcY$, our goal here is to find a good estimator for $\bcX^*$. So the convergence results established in \cite{elden2009newton,savas2010quasi,ishteva2009differential,ishteva2011best} are not directly comparable to ours. Moreover, due to the difference in the targets, the quasi-projection property of T-HOSVD and ST-HOSVD in Tucker tensor decomposition (see Definition \ref{def: quasi-projection Hr}) does not imply they are good estimators for $\bcX^*$. In fact, it has been shown in \cite{zhang2018tensor} that T-HOSVD is strictly suboptimal in estimating $\bcX^*$ in tensor SVD, see more discussions in the second point of Section \ref{sec: result-discussion}.
\end{Remark}

\subsection{A Few More Remarks for Tensor Regression and Tensor SVD} \label{sec: result-discussion}
In this section, we provide a few remarks regarding our results in Theorems \ref{th: tensor regression} and \ref{th: tensor SVD}. 

\begin{itemize}[leftmargin=*]
	\item ({\bf Estimation Error, Convergence Rate, and Signal Strength}) In both tensor regression and SVD, the estimation upper bounds in Theorems \ref{th: tensor regression} and \ref{th: tensor SVD} match the lower bounds in the literature, \cite[Theorem 3]{zhang2018tensor} and \cite[Theorem 5]{zhang2020islet}, which shows that RGN achieves the minimax optimal rate of estimation error. Compared to existing algorithms in the literature on tensor regression and SVD \citep{ahmed2020tensor,chen2016non,han2020optimal,zhang2018tensor}, RGN is the first to achieve the minimax rate-optimal estimation error with only a \emph{double-logarithmic number of iterations} attributed to its second-order convergence.	

Suppose $d,\bar{r}$ are fixed and the condition number $\kappa$ is of order $O(1)$, we note that the sample size requirement ($n \geq O(\sqrt{\bar{r}}\bar{p}^{d/2})$) in tensor regression and least singular value requirement ($\underline{\lambda}/\sigma \geq O(\bar{p}^{d/4}\bar{r}^{1/4})$) in tensor SVD match the start-of-the-arts in literature (see Table \ref{tab: comparison table} for a comparison). Rigorous evidence has been established to show that the least singular value lower bounds in tensor SVD are essential for any polynomial-time algorithm to succeed \citep{zhang2018tensor,brennan2020reducibility,luo2020tensor}. Recent studies in \cite{luo2022tensor,diakonicolas2023statistical} have presented similar evidence for tensor regression, suggesting that a sample complexity of $O(\bar{p}^{d/2})$ is essential for the success of any polynomial-time algorithm.

\item ({\bf Guarantees of the Spectral Initializations in Tensor Regression and Tensor SVD}.) From the proof of Theorem \ref{th: tensor regression} \eqref{ineq: tensor-regression-initialization} and Theorem \ref{th: tensor SVD} \eqref{ineq: tensor-svd-initialization-guarantee}, we show the spectral initializations in tensor regression and tensor SVD have the following guarantees:
	\begin{itemize}[leftmargin=*]
		\item tensor regression:
		\begin{equation} \label{ineq: tensor-regression-init-error-bound}
		\begin{split}
		\|\bcX^0 -\bcX^*\|_{\tHS} \leq C(d) \left( \kappa \widetilde{\sigma} \sqrt{\frac{ \sum_{k=1}^d p_k r_k + \prod_{k=1}^d r_k }{n}} +  \frac{ (\bar{r} \prod_{k=1}^d p_k )^{1/2} \kappa \widetilde{\sigma}^2 }{\underline{\lambda} n}  \right), 
		\end{split}
	\end{equation} where $C(d)$ is some constant depending on $d$ only and $\widetilde{\sigma} :=  \sqrt{\|\bcA\|_{\tHS}^2 + \sigma^2} > \sigma$.
		
		\item tensor SVD:
		\begin{equation} \label{ineq: tensor-svd-initialization-error-bound}
			\begin{split}
				\|\bcX^0 -\bcX^*\|_{\tHS} \leq C(d) \left( \kappa \sigma \sqrt{\sum_{k=1}^d p_k r_k + \prod_{k=1}^d r_k} +  \frac{ (\bar{r} \prod_{k=1}^d p_k )^{1/2} \kappa \sigma }{\underline{\lambda}} \right).
			\end{split}
		\end{equation}
		
	\end{itemize}

Comparing \eqref{ineq: tensor-regression-init-error-bound} and \eqref{ineq: tensor-svd-initialization-error-bound} with \eqref{ineq: tensor-regression-final-error} and \eqref{ineq: tensor-svd-final-error}, we can see the estimation error guarantees of the spectral initializations are strictly suboptimal comparing to the ones with iterative refinement. 
\end{itemize}


\section{Computational Complexity of RGN}\label{sec: computation complexity and implementation}

Next, we investigate the computational complexity of RGN. First, the per-iteration computational cost for RGN with a general linear map $\scA$ is $O(np^dr + n(r^d + dpr)^2)$ if $p_1 = \cdots = p_d = p$ and $r_1 = \cdots = r_d = r$. Here, $O(np^dr)$ and $O(n(r^d + dpr)^2)$ are due to the costs for constructing the covariate maps \eqref{eq: importance sketches} and for solving the least squares problem \eqref{eq: alg least square}, respectively. If $r, d\ll n, p$ (which is a typical case in practice), the cost of constructing the covariates maps dominates and the per-iteration cost of RGN is $O(np^dr)$. 
Performing T-HOSVD or ST-HOSVD in $\cH_{\br}$ can be expensive in general. Since the tensor we apply $\cH_{\br}$ on lies in the tangent space of the current iterate and is at most Tucker rank $2\br$ (Lemma \ref{lm: tangent vector rank 2r property}), the retraction via T-HOSVD and ST-HOSVD can be performed efficiently \citep{cai2020provable}.

A comparison of the per-iteration computational complexity of RGN and several classic algorithms, including alternating minimization (Alter Mini), projected gradient descent (PGD), and gradient descent (GD), in tensor regression and tensor SVD examples is provided in Table \ref{tab: comparison table} in the introduction section. The main complexity of alternating minimization (Alter Mini) \citep{zhou2013tensor,li2013tucker} is from constructing the covariates in solving the least squares and the main complexity of the projected gradient descent (PGD) \citep{chen2016non} and gradient descent (GD) \citep{han2020optimal} are from computing the gradient. We can see RGN has the same per-iteration complexity as Alter Mini and comparable complexity with PGD and GD when $r\ll n, p$. In addition, RGN and Alter Mini are tuning-free, while a proper step size is crucial for PGD and GD to have fast convergence. Finally, RGN enjoys a second-order convergence as shown in Section~\ref{sec:theory}, while the convergence rates of all other algorithms are at most linear. We will further provide a numerical comparison of these algorithms in Section \ref{sec: numerical comparison}.

Furthermore, in specific scenarios where the covariates $\bcA_i$ have more structures, the procedure of RGN can be simplified and the computational complexity can be reduced. In Section \ref{sec:tensor-svd}, we have already seen RGN can be significantly simplified in tensor SVD. Here we discuss two additional scenarios: tensor estimation via rank-1 projections and tensor completion. 
\begin{itemize}[leftmargin=*]
	\item {\bf Tensor estimation via rank-1 projections.} In this application, $\bcA_i = \a_1^{(i)} \circ \cdots \circ \a_d^{(i)}$ and the construction of covariates maps in \eqref{eq: importance sketches} can be simplified as follows
	\begin{equation*}
		(\scA_{\bcB})_i = \U_1^{t\top}\a^{(i)}_{1} \circ \cdots \circ \U_d^{t\top}\a^{(i)}_{d}, \quad (\scA_{\D_k})_{i} = \U_{k\perp}^{t\top} \a_k^{(i)}  (\otimes_{j\neq k} \U_j^{t\top} \a_j^{(i)})^\top \V_k^t, \, k = 1,\ldots, d.
	\end{equation*} 
	Here $\V_k^t = \QR(\cM_k(\bcS^t)^\top)$ and $\bcS^t = \bcX^t \times_{k=1}^d \U_k^{t\top}$. The computational cost for constructing the covariates maps by such a scheme is $O(n(p^2 + r^d))$, which is much cheaper than $O(np^dr)$, the computational cost in the general setting, when $d \geq 3$.
	\item {\bf Tensor completion.} We observe a fraction of entries indexed by $\Omega$ from the target tensor. For $(i_1, \ldots, i_d) \in \Omega$, the corresponding covariate is $\e_{i_1} \circ \cdots \circ \e_{i_d}$. Then, simple calculation yields the covariates maps can be calculated as
	\begin{equation*}
		(\scA_{\bcB})_i = (\U_1^{t\top})_{[:,i_1]} \circ \cdots \circ (\U_d^{t\top})_{[:,i_d]}, \quad (\scA_{\D_k})_{i} = (\U_{k\perp}^{t\top})_{[:,i_k]}  (\otimes_{j\neq k} (\U^t_j)_{[i_j,:]})  \V_k^t, \, k = 1,\ldots, d.
	\end{equation*} 
	The per-iteration cost for constructing the covariates maps above is $O(n(r^d + pr))$.
\end{itemize}

\section{Numerical Studies on RGN}\label{sec:numerics}

We consider four specific numerical settings: 1. tensor regression under random design; 2. tensor estimation via rank-1 projections; 3. tensor completion; 4. tensor SVD. In tensor regression and tensor estimation via rank-1 projections, we generate the covariates $\bcA_i$ and $\a_k^{(i)}$ with i.i.d. $N(0,1)$ entries. In tensor completion, partial observations indexed by $\Omega$ are sampled uniformly at random from the noisy tensor $\bcY$. In each simulation setting, we generate $\bvarepsilon_i \overset{i.i.d.}\sim N(0, \sigma^2)$, $\{\U_k\}_{k=1}^3 $ uniformly at random from $\mathbb{O}_{p, r}$, and $\bcS \in \bbR^{r \times r \times r}$ with i.i.d. $N(0,1)$ entries; then we calculate $\bcX^* = \bcS \times_1 \U_1 \times_2 \U_2 \times_3 \U_3$. In tensor SVD, we also rescale $\bcS$ so that $\min_{k=1,2,3} \sigma_{r_k}(\cM_k(\bcX^*))$ is equal to a pre-specified value $\underline{\lambda}$. 
The implementation details of RGN under each setting have been discussed in Sections \ref{sec: statistics applications} and \ref{sec: computation complexity and implementation}. We apply the following initialization schemes respectively for each problem. 
\begin{itemize}[leftmargin=*]
	\item Tensor regression/Tensor estimation via rank-1 projections: $\bcX^0 = \scA^*(\y) \times_{k=1}^d P_{\U_k^0}$, where $\U_k^0 = \SVD_{r_k}(\cM_k(\scA^*(\y)))$.
	\item Tensor completion: Suppose $\rho = |\Omega|/(\prod_{j=1}^d p_j)$ is the sampling ratio. Denote
	$$\bcY_{\Omega}= \left\{\begin{array}{ll}
	\bcY_{[i_1,\ldots, i_d]}, & \text{if } (i_1,\ldots, i_d)\in \Omega\\
	0, & \text{otherwise}.
	\end{array}\right.$$ 
	Calculate $\mathcal{M}_k(\bcY_{\Omega}) \mathcal{M}_k(\bcY_{\Omega})^\top$, zero out the diagonal entries of $\mathcal{M}_k(\bcY_{\Omega}) \mathcal{M}_k(\bcY_{\Omega})^\top$, let $\U_k^0$ be the leading $r_k$ singular vectors of the diagonal-zero-out matrix $\mathcal{M}_k(\bcY_{\Omega}) \mathcal{M}_k(\bcY_{\Omega})^\top$, and initialize $\bcX^0 = (\bcY_{\Omega}/\rho) \times_{k=1}^d P_{\U_k^0}$ \citep{xia2017statistically}.
	\item Tensor SVD: Initialize $\bcX^0 = \bcY \times_{k=1}^d P_{\U_k^0}$, where $\U_k^0 = \SVD_{r_k}(\cM_k(\bcY))$.
\end{itemize}

Throughout the simulation studies, the error metric we consider is the relative root-mean-squared error (Relative RMSE) $\|\bcX^{t} - \bcX^*\|_{\tHS}/ \|\bcX^*\|_{\tHS}$. The algorithm is terminated when it reaches the maximum number of iterations $t_{\max} = 300$ or the corresponding error metric is less than $10^{-14}$. Unless otherwise noted, the reported results are based on the averages of 50 simulations and on a computer with Intel Xeon E5-2680 2.5GHz CPU. The code of our algorithm can be found at https://github.com/yuetianluo/RGN-for-Tensor-Estimation.

\subsection{Numerical Performance of RGN} \label{sec: numerical RGN property}

\begin{figure}[ht]
  	\centering
  	\subfigure[Tensor estimation via rank-1 projections: $p = 30, r = 3$.]{\includegraphics[width=0.46\textwidth]{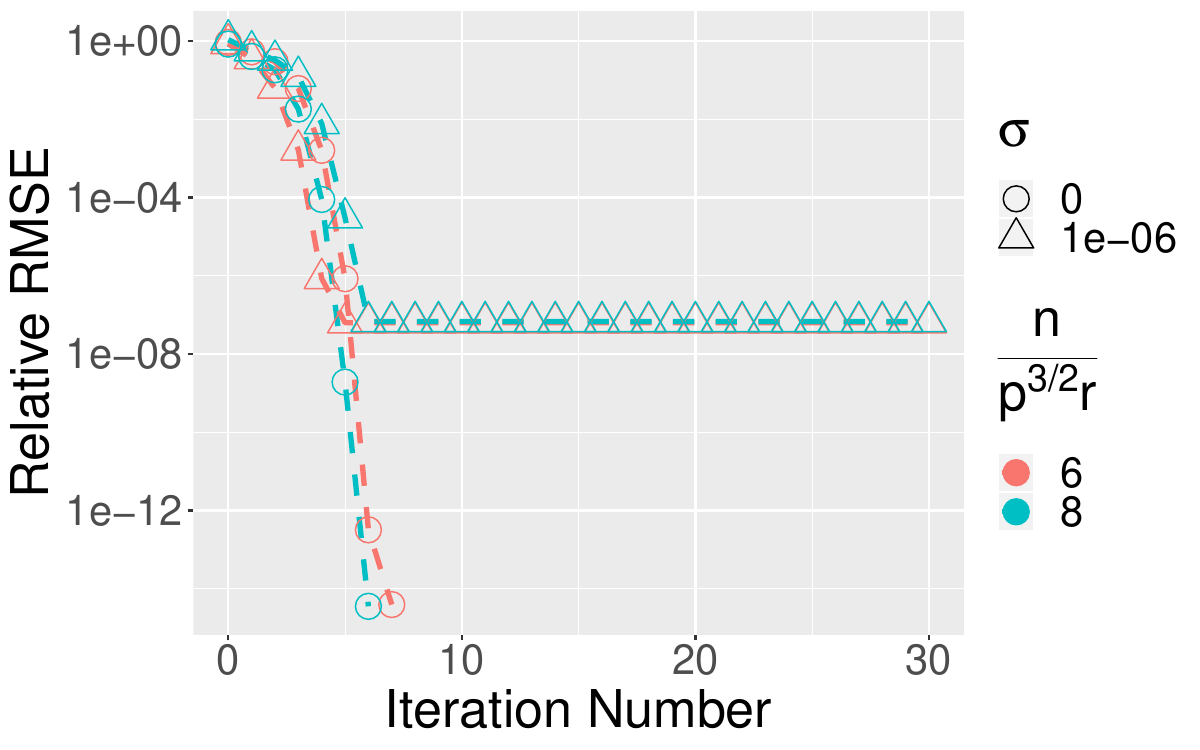}
  	}\qquad 
  	\subfigure[Tensor SVD: $p = 100, r = 3$.]{\includegraphics[width=0.46\textwidth]{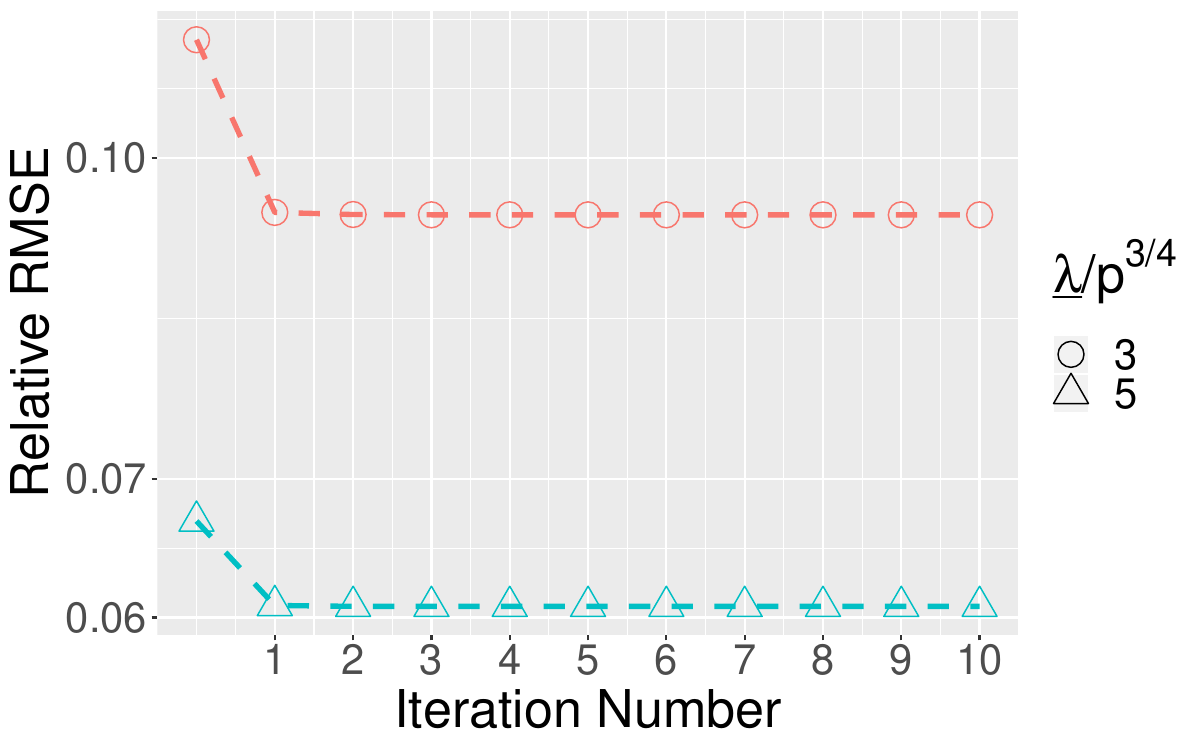}
  	}
  	\caption{Convergence performance of RGN in tensor estimation via rank-1 projections and tensor SVD under spectral initialization.}\label{fig: RGN rankone and decomposition}
\end{figure}

We first examine the convergence rate of RGN in each of the above-mentioned problems. We set $\sigma = 1$ for tensor SVD and $\sigma \in \{0,10^{-6} \}$ in the other three problems. We skip the noiseless setting of tensor SVD since the initialization via T-HOSVD already achieves exact recovery. The convergence performance of RGN in tensor regression and tensor completion is presented in Figure \ref{fig: RGN performance illustration} and the performance in tensor estimation via rank-1 projections and tensor SVD is presented in Figure \ref{fig: RGN rankone and decomposition}. In tensor regression under random design/rank-1 projections and tensor completion, the estimation error converges quadratically to the minimum precision in the noiseless setting and converges quadratically to a limit determined by the noise level in the noisy setting. In tensor SVD, we observe RGN initialized with T-HOSVD converges with almost one iteration. We tried several other simulation settings and observe a similar phenomenon. This suggests that in tensor SVD, RGN may achieve one-step optimality directly after initialization in estimating $\bcX^*$ as discussed in Remark \ref{rem: optimality of RGN}. We leave it as future work to further investigate this phenomenon.

\begin{figure}[h]
  	\centering
  	\subfigure[Tensor regression: $p = 30, r = 3$]{\includegraphics[width=0.46\textwidth]{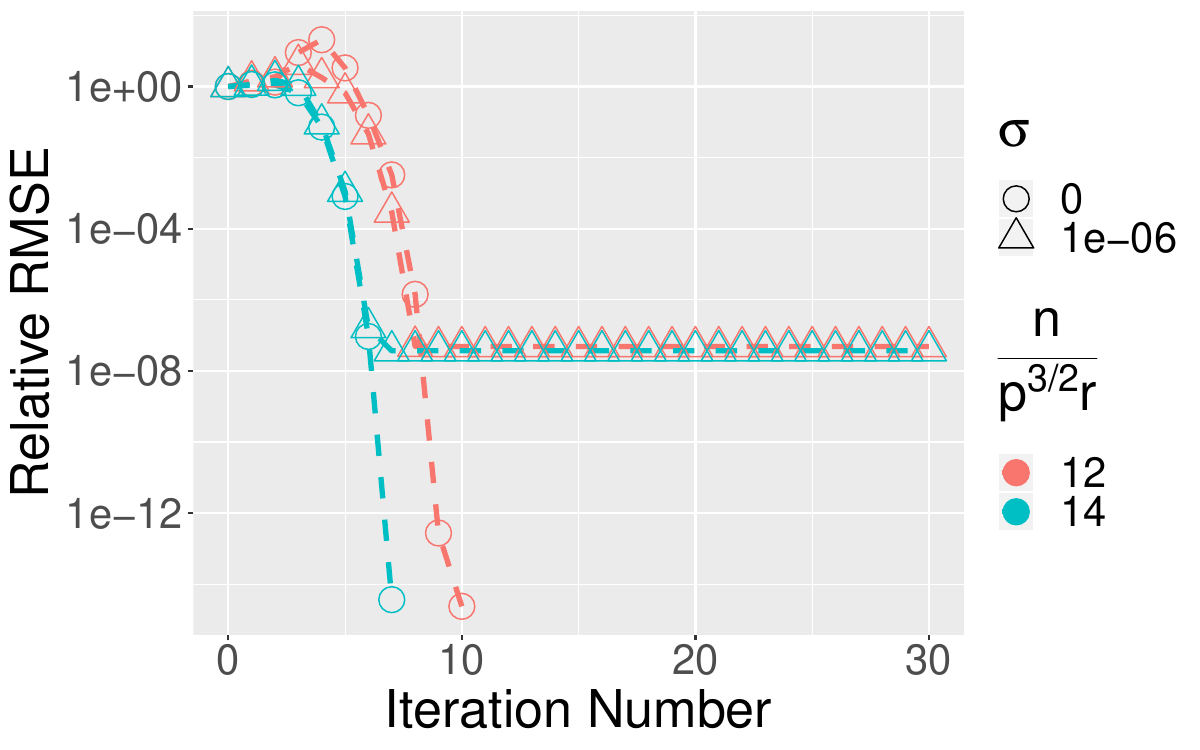}
  	}\qquad 
  	\subfigure[Tensor estimation via rank-1 projections: $p = 30, r = 3$]{\includegraphics[width=0.46\textwidth]{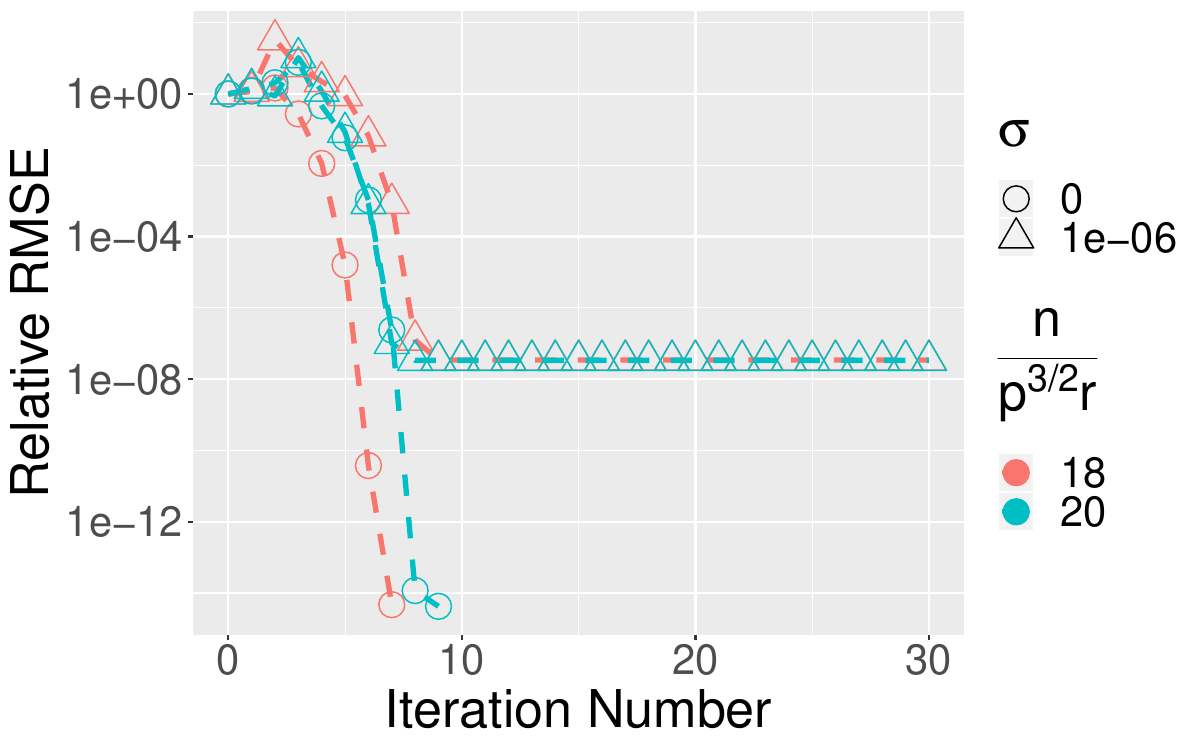}
  	}
  	\caption{Convergence performance of RGN in tensor regression and tensor estimation via rank-1 projections under random initialization}\label{fig: RGN regression and rankone with rand init}
\end{figure}

We note that in many problems, spectral initialization may not be obtainable, and another common initialization choice for iterative algorithms in practice is random initialization. Next, we illustrate the performance of RGN with random initialization in two examples: tensor regression and tensor estimation via rank-1 projections. Here we simply initialize $\bcX^0$ by i.i.d. standard Gaussian entries and the simulation results are given in Figure \ref{fig: RGN regression and rankone with rand init}. We can see that RGN can still converge and estimate/recover $\bcX^*$ well under random initialization. However, there are two main differences compared to the convergence of RGN under spectral initialization: first, we find RGN generally requires a slightly larger sample size to convergence under random initialization; second, the iterate tends to fluctuate at the beginning stage before it enters the attraction region and larger sample size seems to make the algorithm more stable. 

\subsection{Comparison of RGN with Previous Algorithms} \label{sec: numerical comparison}

In this subsection, we compare RGN with other existing algorithms, including the Riemannian trust region method (RTR) \citep{heidel2018riemannian}, alternating minimization (Alter Mini) \citep{zhou2013tensor,li2013tucker} \footnote{Software package available at \cite{zhou2017tensorreg}}, projected gradient descent (PGD)\citep{chen2016non} and gradient descent (GD) \citep{han2020optimal}, in tensor regression. Since the approximate Riemannian Newton in \cite{kressner2016preconditioned} is developed under the setting where the linear map has additive and Kronecker-product-type structures, we choose not to compare it here. While implementing GD and PGD, we evaluate three choices of step size, $\frac{1}{n} *\{0.1,0.5, 1 \}$, then choose the best one following \cite{zheng2015convergent}. We set $p = 30, r = 3, n = 5*p^{3/2}r$ and consider both the noiseless case ($\sigma = 0$) and the noisy case ($\sigma = 10^{-6}$).

\begin{figure}[h]
	\centering
 	\includegraphics[width=0.9\textwidth]{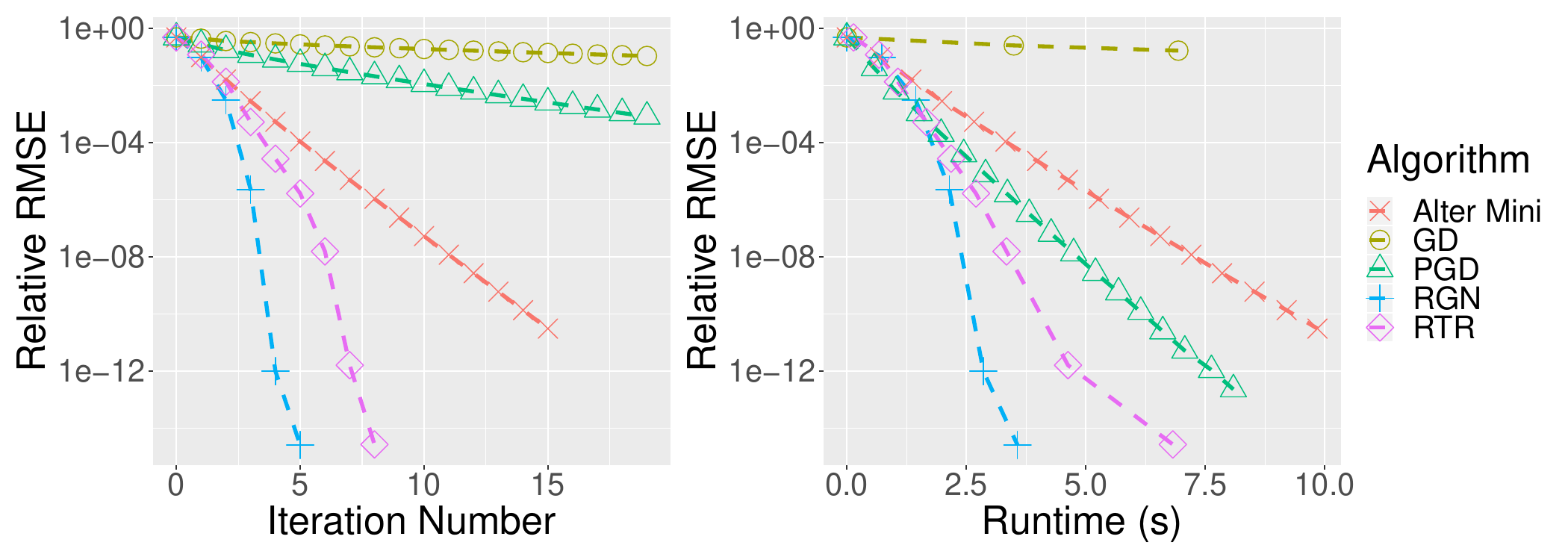}
    \caption{Relative RMSE of RGN (this work), Riemannian trust region (RTR), alternating minimization (Alter Mini), projected gradient descent (PGD), and gradient descent (GD) in noiseless tensor regression. Here, $p = 30, r = 3, n = 5*p^{3/2}r, \sigma = 0$.} \label{fig: tenreg comparison noiseless}
\end{figure}

\begin{figure}[h]
	\centering
	\includegraphics[width=0.9\textwidth]{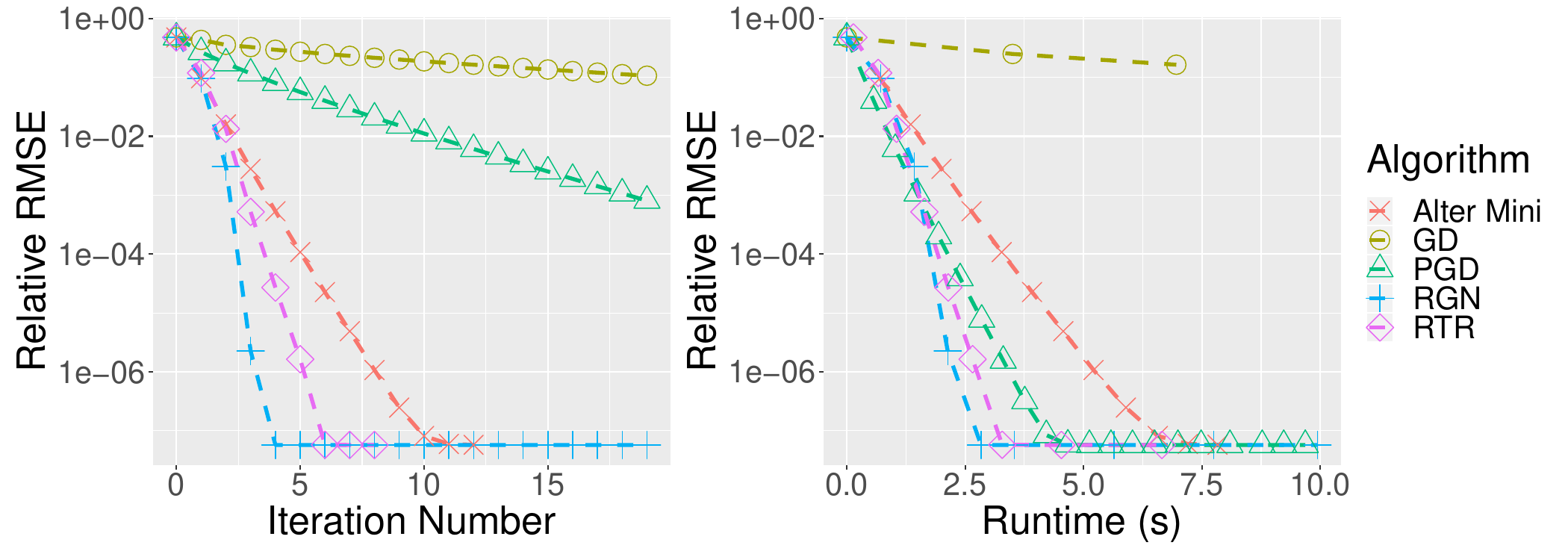}
	\caption{Relative RMSE of RGN (this work), Riemannian trust region (RTR), alternating minimization (Alter Mini), projected gradient descent (PGD) and gradient descent (GD) in noisy tensor regression. Here, $p = 30, r = 3, n = 5*p^{3/2}r, \sigma = 10^{-6}$.} \label{fig: tenreg comparison noisy}
\end{figure}

 We plot the relative RMSE versus iteration number/runtime in both the noiseless and noisy tensor regression settings in Figures \ref{fig: tenreg comparison noiseless} and \ref{fig: tenreg comparison noisy}, respectively. In both settings, RGN converges quadratically, and the Riemannian trust region method is slightly slower than our method but also has superlinear or quadratic convergence performance. All the other baseline algorithms converge in a  much slower linear rate. To achieve an accuracy of $10^{-14}$ in the noiseless setting or the statistical error in the noisy setting, RGN requires a much smaller runtime than PGD, Alter Mini, and GD.

\subsection{A Real Data Example}
\begin{figure}[h]
	\centering
	\includegraphics[width=0.9\textwidth]{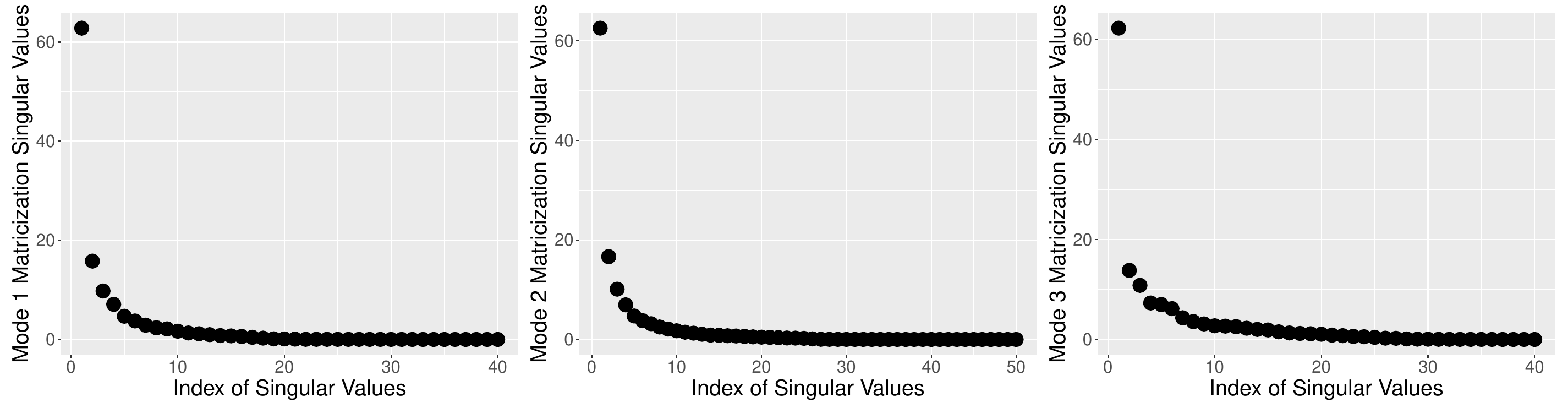}
	\caption{Illustration of low-rankness of $\bcX^*$. Three plots represent the singular values for $\cM_i(\bcX^*)$, $i = 1,2,3$.} \label{fig: decay-singular-value-illustration}
\end{figure}

In this section, we demonstrate the advantages of our algorithm in the application of high-order image compression via rank-1 projection \citep{hao2018sparse,cai2015rop}. We consider the ADHD-200 dataset that contains magnetic resonance imaging (MRI) data from both the attention deficit hyperactivity disorder (ADHD) patients and the control group\footnote{Available at \url{http://neurobureau.projects.nitrc.org/ADHD200/Data.html}}. The dataset includes 973 subjects and each subject is associated with a 121-by-145-by-121 MRI image denoted by $\bcX$. The total storage space for these data through naive format is $121 \times 145 \times 121 \times 973 \times 4\text{B} \approx 7.48 \text{GB}$, which is expensive for both storage and computation. Our goal is to compress the high-order image data via rank-1 projections and allow for efficient retrieval from the compressed rank-1 projections. Here for each image, we choose to let $\bcX^*$ be the subtensor $\bcX_{[1:40,1:50,1:40]}$ with size $40 \times 50 \times 40$. There are two reasons for this choice: first, this experimental scale is already challenging for other algorithms as it is unclear how to leverage the rank-1 projection structure efficiently there; second, since the boundary of an MRI image often contains many zero entries, the selected subtensor has a low-rank structure. An illustration of singular value decay of three matricizations of one MRI image is given in Figure \ref{fig: decay-singular-value-illustration}.

 Specifically, we generate random vectors $\{ \a^{(i)}_1,\a^{(i)}_2, \a^{(i)}_3 \}_{i=1}^n$ with i.i.d. standard Gaussian entries and compute 
\begin{equation*}
	\y_i = \langle \bcX^*, \a_1^{(i)} \circ \a_2^{(i)}  \circ \a_3^{(i)} \rangle, \quad \forall i = 1,\ldots, n.
\end{equation*} 
We can store $\y$ and $\{ \a^{(i)}_1,\a^{(i)}_2, \a^{(i)}_3 \}_{i=1}^n$ in instead of the whole tensor $\bcX^*$. If $n \ll p_1 p_2 p_3/(\max p_i)$, we can reduce the memory cost from $O(p_1 p_2 p_3)$ to $O\left(n(p_1 + p_2 + p_3)\right)$. Furthermore, We can apply our algorithm with inputs $\y$ and $\{ \a^{(i)}_1,\a^{(i)}_2, \a^{(i)}_3 \}_{i=1}^n$ to recover $\bcX^*$. 

\begin{figure}[h]
	\centering
	\includegraphics[width=0.9\textwidth]{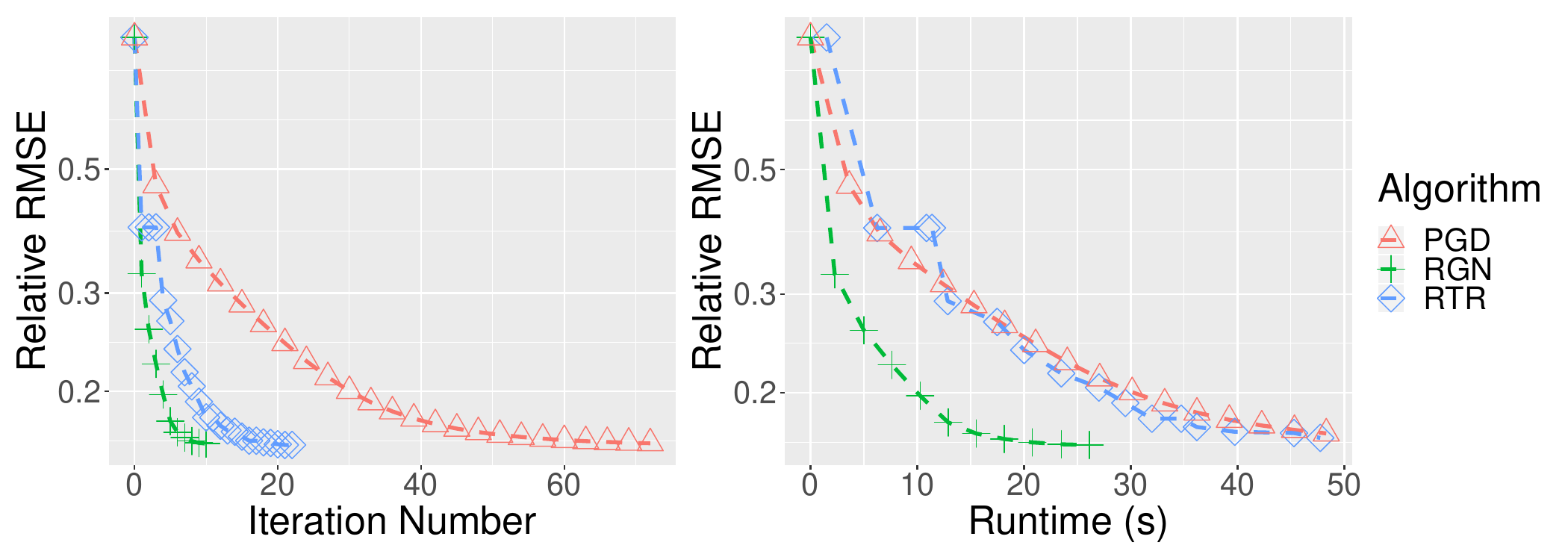}
	\caption{Relative RMSE of RGN (this work), Riemannian trust region (RTR), and projected gradient descent (PGD) over iteration number and runtime in MRI image recovery.} \label{fig: image-recovery-MSE}
\end{figure}

We compare the relative RMSE of our algorithm with the Riemannian trust region method and the PGD method as these two have the best performance from the last simulation study. We stop the algorithm when the decrease of relative RMSE per iteration is less than $10^{-3}$. The recovery performance of these three algorithms with spectral initialization for a randomly drawn MRI image is shown in Figure \ref{fig: image-recovery-MSE}. We can see that in this image all three methods achieve roughly the same relative RMSE, but our algorithm requires significantly less number of iterations and runtime.
 
Moreover, we repeat the experiment for randomly drawn 100 MRI images, and then compare the averaged recovery performance and runtime of these algorithms. The results are given in Table \ref{tab: real-data-comparison-table}. We can see that on average, our method achieves similar recovery guarantees as the Riemannian trust region method but in less runtime. Both these algorithms achieve better recovery performance than PGD. 

\begin{table}[ht]
	\centering
	\begin{tabular}{c |c | c }
	\hline
	\multirow{1}{4em}{Algorithm} & Relative RMSE & Runtime    \\
	\hline
	RGN & $0.155 (0.0137)$ & {\bf 39.2} (6.49)  \\
	\hline
	RTR  & ${\bf 0.153} (0.0137)$ & 104.9 (17.17) \\
	\hline
	PGD &  $0.300 (0.2598)$ & 79.5 (48.79)  \\
	\hline
	\end{tabular}
	\caption{Average relative RMSE and runtime (in second) of  RGN (this work), Riemannian trust region (RTR), and projected gradient descent (PGD) in MRI image recovery. Mean values with standard deviation in the parenthesis are reported.
	}\label{tab: real-data-comparison-table}
\end{table}

\section{Conclusion and Discussions} \label{sec: conclusion}

In this paper, we propose a new algorithm, Riemannian Gauss-Newton (RGN), for low Tucker rank tensor estimation. Under some reasonable assumptions, we show RGN achieves a local quadratic convergence and an optimal statistical error rate for low-rank tensor estimation. 

There are a number of directions worth exploring in the future. First, our current convergence theory relies on the TRIP assumption, which may not hold in scenarios, such as tensor estimation via rank-1 projections and tensor completion. In Section \ref{sec: numerical RGN property}, we show via simulation that RGN still works well without TRIP. It is an interesting future work to establish the ``TRIP-less" theoretical guarantees of RGN. Another future direction is to study the convergence of RGN under random initialization. Some progress has been made on the convergence of randomly initialized (Riemannian) gradient descent in low-rank matrix recovery problems \citep{chen2019gradient,hou2020fast}. However, it can be much harder to establish similar results for RGN in low-rank tensor recovery problems. 

Second, throughout the applications, we assume the noise is Gaussian distributed. In the scenarios that the noise is heavy-tailed or the data have outliers \citep{cai2021generalized}, we would like to consider using the robust loss (e.g., $l_1$ loss or Huber loss) in \eqref{eq: alg least square} instead of the $l_2$ loss or consider quantile tensor regression \citep{lu2020high}. It is interesting to see whether RGN work in those settings and can we give some theoretical guarantees there. 

Third, this paper mainly focuses on the scalar response and tensor predictor model \eqref{eq:model}. In the literature, several papers have also studied the tensor response model \citep{sun2017store,li2017parsimonious}, it is interesting to see whether the RGN method can be applied to that setting. 

Finally, we focus on low Tucker rank tensors in this paper. Although the Tucker format has many advantages, in ultra higher-order tensor problems, the storage cost of the core tensor in the Tucker format scales exponentially with respect to the tensor order and it is more desirable to consider other low-rank tensor decomposition formats, such as the hierarchical Tucker decomposition \citep{grasedyck2010hierarchical,hackbusch2009new} and tensor-train decomposition \citep{oseledets2011tensor,zhou2020optimal}. It is known that the set of fixed hierarchical rank or tensor-train rank tensors forms a smooth manifold \citep{uschmajew2013geometry,hackbusch2012tensor,holtz2012manifolds}, so it is interesting to see whether the RGN algorithms can be established in these settings.


\acks{We thank the editor Suvrit Sra and two anonymous reviewers for their suggestions and comments, which help significantly improve the presentation of this paper. Y. Luo and A. R. Zhang were supported in part by the NSF Grant CAREER-2203741.}


\newpage

\appendix
\section{T-HOSVD and ST-HOSVD} \label{sec: HOSVD, STHOSVD}

In this section, we present the procedures of truncated HOSVD (T-HOSVD) \citep{de2000multilinear} and sequentially truncated HOSVD (ST-HOSVD) \citep{vannieuwenhoven2012new}. For simplicity, we present the sequentially truncated HOSVD with the truncation order from mode $1$ to mode $d$.
\begin{algorithm}[htbp]
	\caption{Truncated High-order Singular Value Decomposition (T-HOSVD)}
	\noindent {\bf Input}: $\bcY \in \bbR^{p_{1} \times \cdots \times p_{d}}$, Tucker rank $\br = (r_1,\ldots, r_d)$.
	\begin{algorithmic}[1]
		\State Compute $\U_k^0 = \SVD_{r_k}(\cM_{k}(\bcY))$ for $k=1, \ldots,d$.
	\end{algorithmic}
	\noindent {\bf Output}: $\widehat{\bcY} = \bcY \times_{k=1}^d P_{\U_k^0}$.
	\label{alg: t-HOSVD}
\end{algorithm}

\begin{algorithm}[htbp]
	\caption{Sequentially Truncated High-order Singular Value Decomposition (ST-HOSVD) }
	\noindent {\bf Input}: $\bcY \in \bbR^{p_{1} \times \cdots \times p_{d}}$, Tucker rank $\br = (r_1,\ldots, r_d)$.
	\begin{algorithmic}[1]
		\State Compute $\U_1^0 = \SVD_{r_1}(\cM_{1}(\bcY))$.
		\For{$k = 2, \ldots, d$}
		\State Compute  $\U_k^0 = \SVD_{r_k}(\cM_{k}(\bcY \times^{k-1}_{l=1} P_{\U_l^0} ))$.
		\EndFor
	\end{algorithmic}
	\noindent {\bf Output}: $\widehat{\bcY} = \bcY \times_{k=1}^d P_{\U_k^0}$.
	\label{alg: st-HOSVD}
\end{algorithm}

\section{Proofs}\label{sec:proof}
We collect all proofs for the main results in this section. We begin by introducing a few preliminary results and then give the proof for all theorems/corollaries/lemmas in subsections.

First, by $\cL_t$ defined in \eqref{eq: Lt}, we can write the least squares in \eqref{eq: alg least square} in the following compact way
\begin{equation} \label{eq: alg compact least square}
		(\bcB^{t+1}, \{\D_k^{t+1}\}_{k=1}^d) = \argmin_{\substack{\bcB\in \bbR^{r_1 \times \cdots \times r_d}, \\ \D_k\in \bbR^{(p_k -r_k) \times r_k}, k=1,\ldots,d}} \left\|\y - \sA \cL_t(\bcB, \{\D_k \}_{k=1}^d )  \right\|_2^2.
\end{equation} 

Also for $\bcX \in \bbM_{\br}$ and the projector $P_{T_\bcX}(\cdot)$ in \eqref{eq: tangent space projector}, we let $P_{(T_{\bcX})_\perp}(\bcZ) := \bcZ - P_{T_{\bcX}}(\bcZ)$ be the orthogonal complement of the projector $P_{T_{\bcX}}$.

\begin{figure}
 		\centering
		\includegraphics[width = 0.4\textwidth]{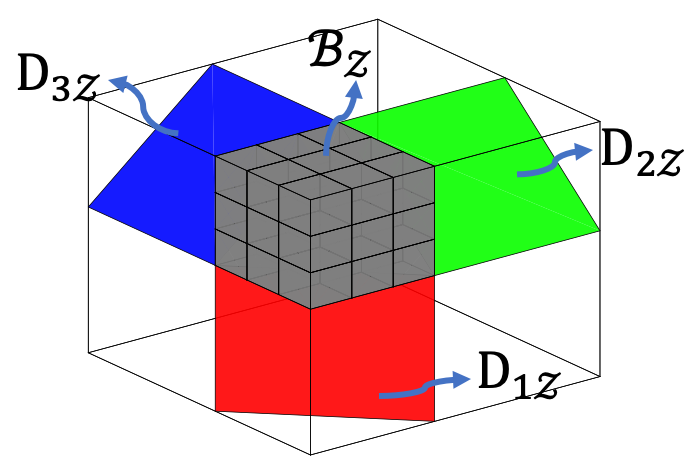}
 		\caption{Illustration of $(\bcB_{\bcZ}, \{\D_{k\bcZ}\}_{k=1}^d)$. Here, we assume $\U_k^\top = [\I_{r_k} ~ \boldsymbol{0}_{r_k \times (p_k-r_k)}]$, $k=1,2,3$, for a better visualization. The gray core tensor represents $\bcB_{\bcZ}$ and red, green, blue blocks represent $\D_{1\bcZ}, \D_{2\bcZ}$ and $\D_{3\bcZ}$, respectively.}\label{fig: illustration of LtZ}
\end{figure}

Next, we introduce a tensorized view of $(\bcZ \times_{k=1}^d \U_k^{t^\top}, \{\U_{k\perp}^{t\top} \cM_k(\bcZ) \W_k^t\}_{k=1}^d)$ generated from $\cL_t^*(\bcZ)$ \eqref{eq: Lt}. For simplicity, denote $\bcB_{\bcZ} = \bcZ \times_{k=1}^d \U_k^{t^\top}$, $\D_{k\bcZ} = \U_{k\perp}^{t\top} \cM_k(\bcZ) \W_k^t$. By construction, it is convenient to view $(\bcB_{\bcZ}, \{\D_{k\bcZ}\}_{k=1}^d)$ lie in a Tucker rank $2\br$ tensor space in $\bbR^{p_1 \times \cdots \times p_d}$ and this fact is useful in the proof. In Figure \ref{fig: illustration of LtZ}, we draw a pictorial illustration to illustrate how does $\cL_t^*(\bcZ)$ look like in a special setting. 

By the explanation above, throughout the proof section, we will use the notation $(\bcB^{t+1}, \{\D_k^{t+1}\}_{k=1}^d)$ to represent a tensor where $\bcB^{t+1}$ and $\{\D_k^{t+1}\}_{k=1}^d$ are located in the same places as $\bcB_{\bcZ}$ and $ \{\D_{k\bcZ}\}_{k=1}^d$ in $\bbR^{p_1 \times \cdots \times p_d}$, and $(\bcB^{t+1}, \{\D_k^{t+1}\}_{k=1}^d) - \cL_t^*(\bcZ)$ denotes the difference of these two tensors.

For any given linear operator ${\cal L}$, we use $\Range(\cL)$ to denote its range space. The first Lemma gives the bounds on the spectrum of $\cL_t^* \scA^* \scA \cL_t$.
\begin{Lemma}[Bounds for Spectrum of $\cL_t^* \scA^* \scA \cL_t$] \label{lm: spectral norm bound of Atop A}
	Recall the definition of $\cL_t$ in \eqref{eq: Lt}. It holds that 
	\begin{equation}
	    \label{eq: norm equal Range Lt}
	    \|\cL_t(\bcZ)\|_{\tHS} = \|\bcZ\|_{\tHS}, \quad \forall \, \bcZ \in \Range(\cL^*_t).
	\end{equation}
	Suppose the linear map $\scA$ satisfies the 2$\br$-TRIP. Then, it holds that for any tensor $\bcZ\in \Range(\cL_t^*)$,
	\begin{equation}\label{eq: spectrum of LAAL}
		(1 - R_{2\br}) \|\bcZ\|_{\tHS} \leq \|\cL_t^* \scA^* \scA \cL_t(\bcZ) \|_{\tHS} \leq (1+R_{2\br})\|\bcZ\|_{\tHS}.
	\end{equation} and
	\begin{equation}\label{eq: spectrum of LAAL inverse}
		\frac{\|\bcZ\|_{\tHS}}{1+R_{2\br}} \leq \|(\cL_t^* \scA^* \scA \cL_t)^{-1}(\bcZ) \|_{\tHS} \leq \frac{\|\bcZ\|_{\tHS}}{1 - R_{2\br}}.
	\end{equation}
\end{Lemma}

{\noindent \bf Proof}. Equation \eqref{eq: norm equal Range Lt} can be directly verified from definitions of $\cL_t$ and $\cL_t^*$ in \eqref{eq: Lt} and the orthogonality for each component in $\cL_t^*$. \eqref{eq: spectrum of LAAL inverse} follows from \eqref{eq: spectrum of LAAL} by the relationship of the spectrum of an operator and its inverse, so we just need to show \eqref{eq: spectrum of LAAL}.

 The second claim is equivalent to say the spectrum of $\cL_t^* \scA^* \scA \cL_t$ is lower and upper bounded by $1 - R_{2\br}$ and $1+R_{2\br}$, respectively, for $\bcZ \in \Range(\cL^*_t)$. Since $\cL_t^* \scA^* \scA \cL_t$ is a symmetric operator, its spectrum can be upper bounded by $\sup_{\bcZ\in \Range(\cL_t^*), \|\bcZ\|_{\tHS} =1 } \langle \bcZ, \cL_t^* \scA^* \scA \cL_t (\bcZ) \rangle $ and lower bounded by $\inf_{\bcZ\in \Range(\cL_t^*), \|\bcZ\|_{\tHS} =1 } \langle \bcZ, \cL_t^* \scA^* \scA \cL_t (\bcZ) \rangle $. Also
\begin{equation*}
\begin{split}
	&\sup_{\bcZ\in \Range(\cL_t^*), \|\bcZ\|_{\tHS} =1 } \langle \bcZ, \cL_t^* \scA^* \scA \cL_t (\bcZ) \rangle = \sup_{\bcZ\in \Range(\cL_t^*), \|\bcZ\|_{\tHS} =1 } \|\scA \cL_t (\bcZ)\|^2_{\tHS} \overset{(a)}\leq 1+R_{2\br}\\
	& \inf_{\bcZ\in \Range(\cL_t^*), \|\bcZ\|_{\tHS} =1 } \langle \bcZ, \cL_t^* \scA^* \scA \cL_t (\bcZ) \rangle = \inf_{\bcZ\in \Range(\cL_t^*), \|\bcZ\|_{\tHS} = 1 } \|\scA \cL_t (\bcZ)\|^2_{\tHS} \overset{(a)}\geq 1-R_{2\br}.
	\end{split}
\end{equation*} Here (a) is by the TRIP condition for $\scA$, $\cL_t(\bcZ)$ is at most Tucker rank $2\br$ and \eqref{eq: norm equal Range Lt}.
 \quad $\blacksquare$

By assuming TRIP for $\scA$, Lemma \ref{lm: spectral norm bound of Atop A} shows the linear operator $\cL_t^* \scA^* \scA \cL_t$ is always invertible over $\Range({\cL}_t^*)$ (i.e. the least squares \eqref{eq: alg least square} has the unique solution). 

Next let us take a detailed look at the error in the least squares in \eqref{eq: alg least square}. To have a better understanding of this least squares, let us rewrite $\y_i$ in the following way
\begin{equation} \label{eq: redecom yi}
	\begin{split}
		\y_i &= \langle \bcA_i, \bcX^* \rangle + \bvarepsilon_i\\
			&= \langle \bcA_i, P_{T_{\bcX^t}} \bcX^* \rangle + \langle \bcA_i, P_{(T_{\bcX^t})_\perp} \bcX^* \rangle + \bvarepsilon_i\\
			& = \underbrace{\langle \cL_t^* \bcA_i, \cL_t^* \bcX^* \rangle}_{(a)} +  \underbrace{\langle \bcA_i, P_{(T_{\bcX^t})_\perp} \bcX^* \rangle + \bvarepsilon_i}_{(b)}\\
			& = \langle \cL_t^* \bcA_i, \cL_t^* \bcX^* \rangle + \bvarepsilon^t_i.
	\end{split}
\end{equation} Here $\bvarepsilon^t := \scA( P_{(T_{\bcX^t})_\perp} \bcX^* ) + \bvarepsilon$. \eqref{eq: redecom yi} can be viewed as the partial linear regression model we considered in performing the least squares in \eqref{eq: alg least square}. In the first expression (a) on the right-hand side of \eqref{eq: redecom yi}, we have the covariates $\cL^*_t(\bcA_i)$ and (b) is the residual in the new partial linear model. Thus, we can see that the estimating target of $(\bcB^{t+1},\{\D_k^{t+1} \}_{k=1}^d)$ is $\cL_t^* \bcX^*$ and its estimation error is given in the following Lemma.
\begin{Lemma}[Least Squares Error in RGN] \label{lm: partial least square for RGN}
Recall the definition of $\bvarepsilon^t = \scA( P_{(T_{\bcX^t})_\perp} \bcX^* ) + \bvarepsilon$ from \eqref{eq: redecom yi}. If the operator $\cL_t^* \scA^* \scA \cL_t$ is invertible over $\Range(\cL_t^*)$, then $(\bcB^{t+1},\{\D_k^{t+1} \}_{k=1}^d)$ in \eqref{eq: alg least square} 
	satisfy	
	\begin{equation}\label{eq: Bt+1 - Btilde}
	\begin{split}
    	(\bcB^{t+1},\{\D_k^{t+1} \}_{k=1}^d) - \cL_t^* \bcX^* = (\cL^*_t \scA^* \scA \cL_t)^{-1} \cL_t^* \scA^* \bvarepsilon^t, 
	\end{split}
\end{equation} and 
	\begin{equation} \label{eq: B and D part bound}
	\|(\bcB^{t+1},\{\D_k^{t+1} \}_{k=1}^d) - \cL_t^* \bcX^* \|_{\tHS} = \left\|(\cL^*_t \scA^* \scA \cL_t)^{-1} \cL_t^* \scA^* \bvarepsilon^t\right\|_{\tHS}.
\end{equation}
\end{Lemma}
{\noindent \bf Proof}. 
First by the decomposition of \eqref{eq: redecom yi}, we have 
\begin{equation}
\label{eq:ydecomp}
\y = \scA \cL_t \cL_t^*(\bcX^*) + \bvarepsilon^t.
\end{equation} In view of \eqref{eq: alg compact least square}, if the operator $\cL_t^* \scA^* \scA \cL_t$ is invertible, the output of least squares in \eqref{eq: alg least square} satisfies
\begin{equation*}
	(\bcB^{t+1},\{\D_k^{t+1} \}_{k=1}^d) = (\cL_t^* \scA^* \scA \cL_t)^{-1} \cL_t^*\scA^* \y = \cL^*_t(\bcX^*) + (\cL_t^* \scA^* \scA \cL_t)^{-1} \cL_t^*\scA^* \bvarepsilon^t,
\end{equation*} 
where the second equality is due to \eqref{eq:ydecomp}. This finishes the proof. \quad $\blacksquare$

Next, we begin the proof for the main results in the paper one by one.

\subsection{Proof of Lemma \ref{lm: mini para of tang space}}
To show the tangent space representation in Lemma \ref{lm: mini para of tang space} is equivalent to the tangent space representation in literature \eqref{eq: old-def-tangent space}, we just need to find a one-to-one correspondence between $\bar{\D}_k$ and $\D_k$ in two representations as the $\bcB$ is the same in both representations. Given $\bar{\D}_k \in \bbR^{p_k \times r_k}$ and $\bar{\D}_k^\top \U_k = \0$, we have $\bar{\D}_k = \U_{k\perp} \M $ for some $\M \in \bbR^{(p_k - r_k) \times r_k}$. So
\begin{equation*}
	\cM_k( \bcS \times_k \bar{\D}_k \times_{i \neq k} \U_i ) \overset{\eqref{eq: matricization relationship}}= \U_{k\perp} \M \cM_k(\bcS) (\otimes_{i = d, i \neq k}^1 \U_i)^\top \overset{(a)}= \U_{k\perp} \M \R^{\top} \W_k^{\top}.   
\end{equation*} Here (a) is because $\V_k=\QR(\cM_k(\bcS)^\top)$ and $\cM_k(\bcS) = \V_k \R$ for some invertible $\R \in \bbR^{r_k \times r_k}$ matrix. Thus, we can see that $\D_k = \M \R^\top$ in the new representation. Similarly, given $\D_k \in \bbR^{(p_k - r_k) \times r_k}$,
\begin{equation*}
	\cT_k(\U_{k\perp} \D_k \W_k^{\top} ) =  \cT_k(\U_{k\perp} \D_k (\R^{\top})^{-1} \R^{\top} \V_k^\top (\otimes_{i = d, i \neq k}^1 \U_i)^\top ) \overset{\eqref{eq: matricization relationship}}= \bcS \times_k \U_{k\perp} \D_k (\R^{\top})^{-1} \times_{i \neq k} \U_i.
\end{equation*} So $\bar{\D}_k = \U_{k\perp} \D_k (\R^{\top})^{-1}$ and this finishes the proof. \quad $\blacksquare$

\subsection{Proof of Lemma \ref{lm: tangent vector rank 2r property}}
Suppose $\bcZ \in T_{\bcX}\bbM_{\br}$ and it has representation $\bcB \times_{k=1}^d \U_k + \sum_{k=1}^d \cT_k(\U_{k\perp}\D_k \W_k^{\top})$ for some $\bcB$ and $\{\D_k\}_{k=1}^d$. To prove the result, it is enough to show $\rank( \cM_k(\bcZ) ) \leq 2r_k$ for $k = 1, \ldots, d$. Let us show this is true for $k = 1$ and the proof for other modes is similar. 

Let $\bcV_k$ be the tensorzied $\V_k^{\top}$ such that $\cM_k(\bcV_k) = \V_k^{\top}$. Recall the definition of $\W_k$ in \eqref{def: Wk}, we have $\cT_k(\U_{k\perp}\D_k \W_k^{\top}) = \bcV_k \times_{l\neq k} \U_l \times_k  \U_{k\perp}\D_k$. So
\begin{equation} \label{eq: tangent 2r check}
    \cM_1(\bcZ) = \U_1 \left( \cM_1(\bcB \times_{k\neq 1} \U_k) + \sum_{k=2}^d \cM_1(\bcV_k \times \U_{k\perp}\D_k \times_{l\neq1,k} \U_l )   \right) + \U_{1\perp}\D_1 \cM_1(\bcV_1 \times_{k\neq 1} \U_k ).
\end{equation} Since $\bcB \in \bbR^{r_1 \times \cdots \times r_d}, \D_1 \in \bbR^{(p_1-r_1) \times r_1}$, each matrix on the right hand side of \eqref{eq: tangent 2r check} is of rank at most $r_1$. Thus $\cM_1(\bcZ)$ is at most rank $2r_1$. This finishes the proof. \quad $\blacksquare$ 

\subsection{Proof of Proposition \ref{prop: Riemannian Gauss-Newton of RGN}}
In view of the Riemannian Gauss-Newton equation in \eqref{eq: Riemannian Gauss-newton equation} and the Riemannian gradient in Lemma \ref{lm:gradient}, to prove the claim, we only need to show
\begin{equation} \label{eq: Thm Rie G-N need to proof}
	P_{T_{\bcX^t}}(\scA^*(\scA( \bcZ^{t+1}) -\y)) = 0.
\end{equation} Here we replace $\eta^{RNG}$ by $\bcZ^{t+1} - \bcX^t$.

From the optimality condition of the least squares problem \eqref{eq: RGN-2}, we know that the least squares solution $\bcZ^{t+1}$ satisfies
\begin{equation}\label{eq:LkZk}
	P_{T_{\bcX^t}}\scA^*\left(\scA P_{T_{\bcX^t}} (\bcZ^{t+1})
 - \y\right) = 0.
\end{equation}
Since $\bcZ^{t+1}$ lies in $T_{\bcX^t} \bbM_{\br}$, $P_{T_{\bcX^t}} (\bcZ^{t+1}) = \bcZ^{t+1}$. Hence, \eqref{eq:LkZk} implies \eqref{eq: Thm Rie G-N need to proof}. \quad $\blacksquare$
\subsection{Proof of Theorem \ref{th: local contraction general setting}}
In this section, we prove our main theorem. This section is divided into two subsections. In the first subsection, we introduce two key Lemmas in proving the results. In the second subsection, we prove the main results. 

\subsubsection{Key Lemmas}
The first Lemma gives an upper bound for the distance of $(\bcB^{t+1},\{\D_k^{t+1} \}_{k=1}^d)$ to their target $\cL_t^* \bcX^*$ in \eqref{eq: B and D part bound}.
\begin{Lemma} [Upper Bound for the Least Squares Estimation Error]\label{lm: bound for iter approx error}
	Let $\bvarepsilon^t = \scA( P_{(T_{\bcX^t})_\perp} \bcX^* ) + \bvarepsilon$. Suppose that $\scA$ satisfies the $3\br$-TRIP. Then at $t$th iteration of RGN, the approximation error \eqref{eq: B and D part bound} has the following upper bound:
	\begin{equation} \label{ineq: prop least square residual bound}
	\begin{split}
		\left\|(\cL^*_t \scA^* \scA \cL_t)^{-1} \cL_t^* \scA^* \bvarepsilon^t\right\|_{\tHS}\leq \frac{R_{3\br} \|P_{(T_{\bcX^t})_\perp} \bcX^*\|_{\tHS} }{1-R_{2\br}} + \frac{\| (\scA^* (\bvarepsilon))_{\max(2\br)}\|_{\tHS}}{1-R_{2\br}}.
	\end{split}
\end{equation}
\end{Lemma}
{\noindent \bf Proof.}  Since $\scA$ satisfies 3$\br$-TRIP, $R_{2\br} \leq R_{3\br} < 1$. Then, Lemma \ref{lm: spectral norm bound of Atop A}'s assumption holds and $\cL^*_t \scA^* \scA \cL_t$ is invertible over $\Range(\cL_t^*)$. 
\begin{equation*}
	\begin{split}
		\left\|(\cL^*_t \scA^* \scA \cL_t)^{-1} \cL_t^* \scA^* \bvarepsilon^t\right\|_{\tHS} & \overset{Lemma\, \ref{lm: spectral norm bound of Atop A}}\leq \frac{1}{1-R_{2\br}} \| \cL_t^* \scA^* \bvarepsilon^t\|_{\tHS}\\
		& \overset{(a)}=  \frac{1}{1-R_{2\br}} \| \cL_t^* \scA^* (\scA( P_{(T_{\bcX^t})_\perp} \bcX^* ) + \bvarepsilon)\|_{\tHS}\\
		& \overset{(b)}\leq \frac{1}{1-R_{2\br}} \left(\| \cL_t^* \scA^* \scA( P_{(T_{\bcX^t})_\perp} \bcX^* ) \|_{\tHS} + \| (\scA^* (\bvarepsilon))_{\max(2\br)}\|_{\tHS} \right)\\
		& \overset{(c)}\leq \frac{R_{3\br} \|P_{(T_{\bcX^t})_\perp} \bcX^*\|_{\tHS} }{1-R_{2\br}} + \frac{\| (\scA^* (\bvarepsilon))_{\max(2\br)}\|_{\tHS}}{1-R_{2\br}},
	\end{split}
\end{equation*} here (a) is by the definition of $\bvarepsilon^t$; (b) is by triangle inequality and $\cL^*_t (\scA^*(\bvarepsilon))$ is of at most Tucker rank $2\br$ as we discussed before; (c) is because
\begin{equation*}
	\begin{split}
		\| \cL_t^* \scA^* \scA( P_{(T_{\bcX^t})_\perp} \bcX^* ) \|_{\tHS} =& \sup_{\bcZ: \|\bcZ\|_{\tHS}\leq 1} \langle \cL_t^* \scA^* \scA( P_{(T_{\bcX^t})_\perp} \bcX^* ), \bcZ \rangle\\
		=& \sup_{\bcZ: \|\bcZ\|_{\tHS}\leq 1} \langle \scA( P_{(T_{\bcX^t})_\perp} \bcX^* ), \scA \cL_t (\bcZ) \rangle\\
		\overset{(a)}\leq & \sup_{\bcZ: \|\bcZ\|_{\tHS}\leq 1}  R_{3\br}\|P_{(T_{\bcX^t})_\perp} \bcX^*\|_{\tHS} \|\cL_t (\bcZ) \|_{\tHS}\\
		\overset{(b)}\leq  & R_{3\br}\|P_{(T_{\bcX^t})_\perp} \bcX^*\|_{\tHS}.
	\end{split}
\end{equation*} Here (a) is due to Lemma \ref{lm:retricted orthogonal property}, $\langle P_{(T_{\bcX^t})_\perp} \bcX^*, \cL_t (\bcZ) \rangle = 0$, $P_{(T_{\bcX^t})_\perp} \bcX^*$ and $\cL_t (\bcZ)$ are of Tucker rank at most $\br$ and $2\br$, respectively; (b) is because $\|\cL_t (\bcZ) \|_{\tHS} \leq \|\bcZ\|_{\tHS} \leq 1$.  
 \quad $\blacksquare$

The following Lemma plays a key role in showing the quadratic convergence of RGN.
\begin{Lemma}[Projection of $\bcX^*$ on $P_{(T_{\bcX^t})_\perp}$ ] \label{lm: orthogonal projection}
	For any two order-$d$ Tucker rank $\br := (r_1, \ldots,r_d)$ tensors $\bcX^*, \bcX^t \in \bbR^{p_1 \times \cdots \times p_d}$, we have
	\begin{equation*}
		\| P_{(T_{\bcX^t})_\perp} \bcX^* \|_{\tHS} \leq \frac{d \| \bcX^t - \bcX^* \|^2_{\tHS} }{\underline{\lambda}},
	\end{equation*} where $\underline{\lambda} := \min_{k=1,\ldots,d} \sigma_{r_k}(\cM_k(\bcX^*))$.
\end{Lemma}

{\noindent \bf Proof}. Suppose $\bcX^t$ and $\bcX^*$ have Tucker rank $\br$ decomposition $\llbracket \bcS^t; \U_1^t,\ldots,\U_d^t \rrbracket $ and $\llbracket \bcS; \U_1,\ldots,\U_d \rrbracket $, respectively. Recall 
\begin{equation*} 
	\W_k  := (\U_d \otimes \cdots \otimes \U_{k+1} \otimes \U_{k-1} \otimes \ldots \otimes \U_1) \V_k \in \bbO_{p_{-k}, r_{k}}
\end{equation*} in \eqref{def: Wk}, where $\V_k = \QR(\cM_k(\bcS)^\top)$. Similarly we have $\W_k^t$ for $\bcX^t$. For $\bcX^*$, it can be decomposed in the following way
\begin{equation} \label{eq: decom of Xstar}
\begin{split}
	\bcX^* = & \bcX^* \times_1 P_{\U_{1\perp}^t} + \bcX^* \times_1 P_{\U_{1}^t} \times_2 P_{\U_{2\perp}^t} + \cdots +  \bcX^* \times_{l=1}^{k-1} P_{\U_{l}^t} \times_k P_{\U_{k\perp}^t} + \cdots + \bcX^* \times_{l=1}^{d} P_{\U_{l}^t}\\
		=& \sum_{k=1}^d \bcX^* \times_{l=1}^{k-1} P_{\U_{l}^t} \times_k P_{\U_{k\perp}^t} + \bcX^* \times_{l=1}^{d} P_{\U_{l}^t}
\end{split}
\end{equation} Then
\begin{equation} \label{eq: P-Tperp Xstar}
	\begin{split}
		P_{(T_{\bcX^t})_\perp} \bcX^* =& \bcX^* - P_{T_{\bcX^t}} \bcX^*\\
	\overset{\eqref{eq: tangent space projector}}=	& \bcX^* - (\bcX^* \times_{k=1}^d P_{\U_k^t} + \sum_{k=1}^d \cT_k( P_{\U_{k\perp}^t} \cM_k(\bcX^*) P_{\W_k^t} )  )\\
	\overset{\eqref{eq: decom of Xstar}}= & \sum_{k=1}^d \left(\bcX^* \times_{l=1}^{k-1} P_{\U_{l}^t} \times_k P_{\U_{k\perp}^t} - \cT_k( P_{\U_{k\perp}^t} \cM_k(\bcX^*) P_{\W_k^t} ) \right)\\
	\overset{ \eqref{eq: matricization relationship} }= &  \sum_{k=1}^d \left( \cT_k \left( P_{\U_{k\perp}^t} \cM_k(\bcX^*) ( \otimes_{l=d}^{k+1}\I_{p_l}\otimes_{l=k-1}^{1} P_{\U_l^t} - P_{\W_k^t}  )   \right)  \right) \\
	\overset{(a)} = & \sum_{k=1}^d \left( \cT_k \left( (P_{\U_k} - P_{\U_{k}^t}) \cM_k(\bcX^*) ( \otimes_{l=d}^{k+1}\I_{p_l}\otimes_{l=k-1}^{1} P_{\U_l^t} - P_{\W_k^t}  )   \right)  \right)\\
	\overset{(b)} = & \sum_{k=1}^d \left( \cT_k \left( (P_{\U_k} - P_{\U_{k}^t}) \cM_k(\bcX^* - \bcX^t) ( \otimes_{l=d}^{k+1}\I_{p_l}\otimes_{l=k-1}^{1} P_{\U_l^t} - P_{\W_k^t}  )   \right)  \right),
	\end{split}
\end{equation} here (a) is because the $\U_k$ spans the column space of $\cM_k(\bcX^*)$, (b) is because $\cM_k(\bcX^t) ( \otimes_{l=d}^{k+1}\I_{p_l}\otimes_{l=k-1}^{1} P_{\U_l^t} - P_{\W_k^t} ) = 0$.

It is easy to check $\otimes_{l=d}^{k+1}\I_{p_l}\otimes_{l=k-1}^{1} P_{\U_l^t} - P_{\W_k^t}$ is a projection matrix. So from \eqref{eq: P-Tperp Xstar}, we have
\begin{equation*}
	\begin{split}
		\|P_{(T_{\bcX^t})_\perp} \bcX^*\|_{\tHS}\leq &  \sum_{k=1}^d \| \cT_k \left( (P_{\U_k} - P_{\U_{k}^t}) \cM_k(\bcX^* - \bcX^t) ( \otimes_{l=d}^{k+1}\I_{p_l}\otimes_{l=k-1}^{1} P_{\U_l^t} - P_{\W_k^t}  ) \right)\|_{\tHS}\\
		\leq &\sum_{k=1}^d \|(\bcX^* - \bcX^t) \times_k (P_{\U_k} - P_{\U_{k}^t} ) \|_{\tHS}\\
		\leq &  d\|\bcX^* - \bcX^t\|_{\tHS} \max_{k=1,\ldots,d} \|P_{\U_k} - P_{\U_{k}^t}\|\\
		\overset{(a)} \leq  & \frac{d \| \bcX^t - \bcX^* \|^2_{\tHS}}{\underline{\lambda}},
	\end{split} 
\end{equation*} here (a) is due to the matrix subspace perturbation bound $\|P_{\U_k} - P_{\U_{k}^t}\| \leq \frac{\|\cM_k(\bcX^*) - \cM_k(\bcX^t)\|}{\sigma_{r_k}(\cM_k(\bcX^*))} \leq \frac{\|\cM_k(\bcX^*) - \cM_k(\bcX^t)\|_F}{\sigma_{r_k}(\cM_k(\bcX^*))} = \frac{\|\bcX^* - \bcX^t\|_{\tHS}}{\sigma_{r_k}(\cM_k(\bcX^*))}$ (for example see Lemma 4.2 of \cite{wei2016guarantees}). This finishes the proof of this Lemma.
\quad $\blacksquare$

We note a similar result to Lemma \ref{lm: orthogonal projection} appears in Lemma 5.2 of \cite{cai2020provable} and here we have exponential improvement on the dependence of $d$.
\subsubsection{Proof of Theorem \ref{th: local contraction general setting}}
Now we prove the main results. First, notice the convergence result in the noiseless setting follows easily from the noisy setting by setting $\bvarepsilon = 0$. Suppose $\cH_{\br}$ satisfies the quasi-projection property with approximation constant $\delta(d)$. For notation simplicity, let us denote $\bcX^{t+0.5} := \bcZ^{t+1} = \cL_t(\bcB^{t+1},\{\D_k^{t+1} \}_{k=1}^d) = \bcB^{t+1} \times_{k=1}^d \U^t_k + \sum_{k=1}^d \cT_k(\U_{k\perp}^{t}\D_k^{t+1}\W_k^{t\top})$.
\begin{equation} \label{ineq: Xt+1 -Xstar}
	\begin{split}
		\|\bcX^{t+1} - \bcX^*\|_{\tHS} =& \| \cH_{\br}(\bcX^{t+0.5}) - \bcX^* \|_{\tHS}\\
		 \leq & \| \cH_{\br}(\bcX^{t+0.5}) - \bcX^{t+0.5}\|_{\tHS} + \|\bcX^{t+0.5} - \bcX^*\|_{\tHS} \\
		\overset{(a)}\leq & \delta(d) \|P_{\bbM_{\br}}(\bcX^{t+0.5}) - \bcX^{t+0.5}\|_{\tHS} + \|\bcX^{t+0.5} - \bcX^*\|_{\tHS}\\
		\overset{(b)}\leq & (\delta(d) + 1)\|\bcX^{t+0.5} - \bcX^*\|_{\tHS}\\
		=  & (\delta(d) + 1)\|\cL_t(\bcB^{t+1},\{\D_k^{t+1} \}_{k=1}^d) - P_{T_{\bcX^t}}\bcX^* - P_{(T_{\bcX^t})_\perp}\bcX^*\|_{\tHS}\\
		\overset{(c)}\leq & (\delta(d) + 1) \left(\|\cL_t(\bcB^{t+1},\{\D_k^{t+1} \}_{k=1}^d) - P_{T_{\bcX^t}}\bcX^*\|_{\tHS} + \|P_{(T_{\bcX^t})_\perp}\bcX^*\|_{\tHS} \right)\\
		\overset{Lemma\, \ref{lm: spectral norm bound of Atop A} }= & (\delta(d) + 1) \left(\|(\bcB^{t+1},\{\D_k^{t+1} \}_{k=1}^d) - \cL_t^*(\bcX^*)\|_{\tHS} + \|P_{(T_{\bcX^t})_\perp}\bcX^*\|_{\tHS} \right)
	\end{split}
\end{equation} here (a) is by the quasi-projection property of $\cH_{\br}$; (b) is by the projection property of $P_{\bbM_{\br}}(\cdot)$; (c) is by triangle inequality.

Notice that by the $3\br$-TRIP assumption of $\scA$, the assumptions in Lemma \ref{lm: partial least square for RGN} and \ref{lm: bound for iter approx error} are satisfied. By Lemma \ref{lm: partial least square for RGN}, \ref{lm: bound for iter approx error} and \ref{lm: orthogonal projection}, the right hand side of \eqref{ineq: Xt+1 -Xstar} can be bounded as follows
\begin{equation*}
	\begin{split}
		\|\bcX^{t+1} - \bcX^*\|_{\tHS} \overset{Lemma\, \ref{lm: partial least square for RGN}, \ref{lm: bound for iter approx error}} \leq & (\delta(d)+1)\left( (1 +\frac{R_{3\br} }{1-R_{2\br}} ) \|P_{(T_{\bcX^t})_\perp}\bcX^*\|_{\tHS} +  \frac{\| (\scA^* (\bvarepsilon))_{\max(2\br)}\|_{\tHS}}{1-R_{2\br}} \right)\\
		\overset{Lemma\, \ref{lm: orthogonal projection} }\leq &  d(\delta(d) + 1)(\frac{R_{3\br}}{1-R_{2\br}} + 1) \frac{\|\bcX^t - \bcX^*\|^2_{\tHS}}{\underline{\lambda}} + \frac{\delta(d) + 1}{1 - R_{2\br}} \|(\scA^*(\bvarepsilon ))_{\max(2\br)}\|_{\tHS}.
	\end{split}
\end{equation*} Finally, the convergence is guaranteed under the initialization condition. By taking $\delta(d) = \sqrt{d}$, we finish the proof of this Theorem.  \quad $\blacksquare$

\subsection{Proof of Corollary \ref{coro: two-phase convergence}}
Recall from Theorem \ref{th: local contraction general setting}, we have 
\begin{equation*}
    \|\bcX^{t+1} - \bcX^* \|_{\tHS} \leq \underbrace{d(\delta(d) + 1)(\frac{R_{3\br}}{1-R_{2\br}} + 1) \frac{\|\bcX^t - \bcX^*\|^2_{\tHS}}{\underline{\lambda}}}_{(A1)} + \underbrace{\frac{\delta(d) + 1}{1 - R_{2\br}} \|(\scA^*(\bvarepsilon ))_{\max(2\br)}\|_{\tHS}}_{(A2)}.
\end{equation*} 
Notice, in the first phase, (A1) dominates (A2) and in the second phase (A2) dominates (A1) and the two-phase convergence results follow. 

By induction, it is easy to show under the initialization condition, in the first phase we have
\begin{equation*}
	\|\bcX^{t} - \bcX^* \|_{\tHS} \leq 2^{-2^{t}} \|\bcX^{0} - \bcX^*\|_{\tHS}. 
\end{equation*}

When $t \geq T_{\max}$ indicated in the Theorem, the algorithm enters the second phase and $ \|\bcX^{t} - \bcX^* \|_{\tHS} \leq \frac{2(\delta(d) + 1)}{1 - R_{2\br}} \|(\scA^*(\bvarepsilon ))_{\max(2\br)}\|_{\tHS}$. Combining phases 1 and 2, we have
\begin{equation*}
	\|\bcX^{t} - \bcX^* \|_{\tHS} \leq 2^{-2^{t}} \|\bcX^{0} - \bcX^*\|_{\tHS} + \frac{2(\delta(d) + 1)}{1 - R_{2\br}} \|(\scA^*(\bvarepsilon ))_{\max(2\br)}\|_{\tHS}
	\end{equation*}
for all $t$. By taking $\delta(d) = \sqrt{d}$, we finish the proof. \quad $\blacksquare$
\subsection{Proof of Theorem \ref{th: pertur lower bound}}
The proof is done by construction. We consider a very special setting where $\scA(\bcZ) = \rmvec(\bcZ)$, and here $\rmvec(\bcZ)$ denotes the vectorization of $\bcZ$. This can be viewed as the tensor decomposition setting and we can tensorize model \eqref{eq:model} and get $\bcY = \bcX^* + \bcE$. It is easy to verify $\scA$ satisfies the TRIP condition and here $\xi = \|\bcE_{\max(2\br)}\|_{\tHS}$. Let us denote $r = \min_{k=1,\ldots,d} r_k$ and $\bcI_r \in \bbR^{r \times r \times \cdots \times r}$ as the order-$d$ identity tensor with entries $(i,i,\ldots, i)$ to be $1$ and others are $0$. We construct 
\begin{equation*}
\bcE_1 = \frac{\xi}{\sqrt{r}} \bcI_r \times_1 \left(  \begin{array}{c}
\0_{r \times r}  \\
\I_r \\
\0_{(p_1-2r) \times r}
\end{array} \right) \times \cdots \times_d \left(  \begin{array}{c}
\0_{r \times r}  \\
\I_r \\
\0_{(p_d-2r) \times r}
\end{array} \right),
\end{equation*}where $\0_{m \times n}$ denotes a $m \times n$ matrix with all entries to be $0$. 

It is easy to check that $ \|(\bcE_1)_{\max(2\br)}\|_{\tHS} = \xi $. Similarly, we construct 
\begin{equation*}
\bcX_1 = \frac{\xi}{\sqrt{r}} \bcI_r \times_1 \left(  \begin{array}{c}
\I_r \\
\0_{r \times r} \\
\0_{(p_1-2r) \times r}
\end{array} \right) \times \cdots \times_d \left(  \begin{array}{c}
\I_r  \\
\0_{r \times r} \\
\0_{(p_d-2r) \times r}
\end{array} \right).
\end{equation*}

Also we let $\bcE_2 = \bcX_1$ and $\bcX_2 = \bcE_1$, and it is easy to check $(\scA, \rmvec(\bcX_1),\rmvec(\bcE_1) )\in \mathcal{F}_{\br}(\xi)$ and $ (\scA, \rmvec(\bcX_2), \rmvec(\bcE_2) ) \in \mathcal{F}_{\br}(\xi)$. At the same time, we have $\bcE_1 + \bcX_1 = \bcX_2 + \bcE_2$. Thus
\begin{equation*}
\begin{split}
\inf_{\widehat{\bcX}} \sup_{(\widetilde{\scA},\widetilde{\bcX},\widetilde{\bvarepsilon}) \in \mathcal{F}_{\br}(\xi)} \| \widehat{\bcX} - \widetilde{\bcX} \|_{\tHS} &\geq \inf_{\widehat{\bcX}} \max \left\{ \| \widehat{\bcX} - \bcX_1 \|_{\tHS}, \| \widehat{\bcX} - \bcX_2 \|_{\tHS}  \right\} \\
& \geq \frac{1}{2} \left(\| \widehat{\bcX} - \bcX_1 \|_{\tHS} + \| \widehat{\bcX} - \bcX_2 \|_{\tHS}\right)\\
& \geq \frac{1}{2} \|\bcX_1 - \bcX_2\|_{\tHS} = \frac{\sqrt{2}}{2} \xi.  
\end{split}
\end{equation*}
\quad $\blacksquare$

\subsection{Proof of Lemma \ref{lm: RGN in ten-decom}}

Note that in the tensor decomposition model, the map $\scA$ can be viewed as an identity map, and we have the tensorized model $\bcY = \bcX^* + \bcE$. The objective function of the least squares in \eqref{eq: alg least square} can be written in the following way
\begin{equation} \label{eq: leasts square obj}
	\begin{split}
		&\| \bcY - \bcB \times_{k=1}^d \U_k^t - \sum_{k=1}^d \cT_k ( \U_{k\perp}^t \D_k \W_k^{t\top} ) \|_{\tHS}^2\\
		= & \| \bcY - \cL_t( \bcB, \{\D_k \}_{k=1}^d ) \|_{\tHS}^2\\
		\overset{(a)}= & \| P_{T_{\bcX^t}}\bcY - \cL_t( \bcB, \{\D_k \}_{k=1}^d ) \|_{\tHS}^2 + \|P_{(T_{\bcX^t})_\perp}\bcY\|^2_{\tHS} \\
		=& \|\cL_t (\cL^*_t(\bcY) - (\bcB, \{\D_k \}_{k=1}^d )) \|_{\tHS}^2 + \|P_{(T_{\bcX^t})_\perp}\bcY\|^2_{\tHS}
	\end{split}
\end{equation} here (a) is because $P_{T_{\bcX^t}}\bcY$ and $\cL_t( \bcB, \{\D_k \}_{k=1}^d )$ lie in the tangent space of $\bcX^t$, while $P_{(T_{\bcX^t})_\perp}\bcY$ lies in the orthogonal complement of the tangent space of $\bcX^t$.

Thus, to minimize the loss function, we just need to minimize the first term on the right-hand side of \eqref{eq: leasts square obj} and it is easy to see the minimizer is obtained when $(\bcB^{t+1}, \{\D^{t+1}_k \}_{k=1}^d)$ is equal to $\cL^*_t(\bcY)$. The results follow by the definition of $\cL_t^*$ in \eqref{eq: Lt}. \quad $\blacksquare$

\subsection{Proof of Theorem \ref{th: tensor regression} }
First the $3\br$-TRIP condition for $\scA$ holds when $n \geq C( \bar{r}^d + d \bar{p} \bar{r} )/R^2_{3\br}$ by \cite[Theorem 2]{rauhut2017low}. Denote $\widetilde{\sigma}^2 =  \|\bcA\|_{\tHS}^2 + \sigma^2$, we claim with probability at least $1-cc\underline{p}^{-C}$, the following inequalities hold:
\begin{equation} \label{ineq: tensor regression two fact}
\begin{split}
	&\left\|\sin\Theta(\U_k^0,\U_k)\right\|= \|\U_{k\perp}^{0\top} \U_k\| \leq \frac{\sqrt{p_k/n}\widetilde{\sigma} \underline{\lambda} + (\prod_{k=1}^d p_k)^{1/2} \widetilde{\sigma}^2 /n}{\underline{\lambda}^2} \\
	& \|(\scA^*(\bvarepsilon))_{\max(2\br)}\|_{\tHS} \leq C\sigma\sqrt{\frac{\sum_{k=1}^d r_k p_k + \prod_{k=1}^d r_k}{n} }.
\end{split}
\end{equation} Here the first result holds when $n \geq c(d) (\|\bcX^*\|_{\tHS}^2 + \sigma^2)  \frac{ \bar{p}^{d/2} }{\underline{\lambda}^2}$ by the proof of Theorem 4 in \cite{zhang2020islet} and the second result is by Lemma \ref{lm: concentration of noise}.

Since $\bcX^0 = (\scA^*(\scA(\bcX^*) +  \bvarepsilon )) \times_{k=1}^d P_{\U_k^0}$, by Lemma \ref{lm: tensor estimation via projection}, we have
\begin{equation*} 
	\left\|\bcX^0 - \bcX^*\right\|_{\tHS} \leq \left\| (\scA^* \scA (\bcX^*) - \bcX^*  + \scA^*(\bvarepsilon) )   \times_{k=1}^d P_{\U^0_k} \right\|_{\tHS} + \sum_{k=1}^d \left\| \U_{k\perp}^{0\top} \cM_k(\bcX^*) \right\|_F. 
\end{equation*}

For $k=1,\ldots,d$
\begin{equation*} 
	\begin{split}
		\left\| \U_k^{0\top} \cM_k(\bcX^*) \right\|_F =& \|\U_{k\perp}^{0\top} \U_k \U_k^\top \cM_k(\bcX^*)  \|_F\\
		\leq & \|\U_{k\perp}^{0\top} \U_k\|_F \|\U_k^\top \cM_k(\bcX^*)\|\\
		\overset{\eqref{ineq: tensor SVD two fact} } \leq &\frac{\sqrt{p_k r_k/n}\widetilde{\sigma} \underline{\lambda} + (r_k \prod_{k=1}^d p_k)^{1/2} \widetilde{\sigma}^2 /n}{\underline{\lambda}^2} \bar{\lambda}.
	\end{split}
\end{equation*}

Moreover, 
\begin{equation} \label{ineq: the-first-term-bound}
	\begin{split}
		 &\left\| (\scA^* \scA (\bcX^*) - \bcX^*  + \scA^*(\bvarepsilon) )   \times_{k=1}^d P_{\U^0_k} \right\|_{\tHS} \\
		 \leq & \| (\scA^* \scA (\bcX^*) - \bcX^*)   \times_{k=1}^d P_{\U^0_k} \|_{\tHS} + \|\scA^*(\bvarepsilon) \times_{k=1}^d P_{\U^0_k} \|_{\tHS}.
	\end{split}
\end{equation}

Notice that by Lemma \ref{lm: concentration of noise}, with probability at least $1 - \exp(-C\underline{p})$, we have
\begin{equation} \label{ineq: tensor-regression-bound-part1}
	\|\scA^*(\bvarepsilon) \times_{k=1}^d P_{\U^0_k} \|_{\tHS} \leq \left\| (\scA^*(\bvarepsilon ))_{\max(2\br)} \right\|_{\tHS} \leq C'\sigma\sqrt{\frac{\sum_{k=1}^d r_k p_k + \prod_{k=1}^d r_k}{n} }.
\end{equation}
In addition
\begin{equation}\label{ineq: tensor-regression-bound-part2}
	\begin{split}
		\| (\scA^* \scA (\bcX^*) - \bcX^*)   \times_{k=1}^d P_{\U^0_k} \|_{\tHS} &= \sup_{\bcZ: \|\bcZ\|_{\tHS} = 1} \langle (\scA^* \scA (\bcX^*) - \bcX^*)   \times_{k=1}^d P_{\U^0_k}, \bcZ \rangle \\
		& = \sup_{\bcZ: \|\bcZ\|_{\tHS} = 1} \langle (\scA^* \scA (\bcX^*) - \bcX^*) , \bcZ \times_{k=1}^d P_{\U^0_k} \rangle \\
		& \leq  \sup_{\bcZ: \|\bcZ\|_{\tHS} = 1} | \langle \scA(\bcX^*), \scA(\bcZ \times_{k=1}^d P_{\U^0_k}) \rangle - \langle \bcX^*, \bcZ \times_{k=1}^d P_{\U^0_k} \rangle | \\
		& \overset{ \text{Lemma } \ref{lm:retricted orthogonal property} } \leq \sup_{\bcZ: \|\bcZ\|_{\tHS} = 1} R_{2\br} \|\bcX^*\|_{\tHS}  \|\bcZ \times_{k=1}^d P_{\U^0_k}\|_{\tHS} \\
		& \leq R_{2\br} \|\bcX^*\|_{\tHS} \leq C\sqrt{\frac{ \sum_{k=1}^d r_k p_k + \prod_{k=1}^d r_k}{n}} \|\bcX^*\|_{\tHS},
	\end{split}
\end{equation} where the last inequality uses the fact $R_{2\br}$ is of order $\sqrt{\frac{ \sum_{k=1}^d r_k p_k + \prod_{k=1}^d r_k}{n}}$ by \cite[Theorem 2]{rauhut2017low}.

By plugging \eqref{ineq: tensor-regression-bound-part1}, \eqref{ineq: tensor-regression-bound-part2} into \eqref{ineq: the-first-term-bound}, we have
\begin{equation} \label{ineq: tensor-regression-initialization}
\begin{split}
	\|\bcX^0 -\bcX^*\|_{\tHS} & \leq C(d) \left( \kappa \sum_{k=1}^d \sqrt{\frac{p_k r_k}{n}}\widetilde{\sigma} + \sqrt{\frac{\prod_{k=1}^d r_k}{n}}(\sigma + \|\bcX^*\|_{\tHS} ) + \frac{ (\bar{r} \prod_{k=1}^d p_k )^{1/2} \kappa \widetilde{\sigma}^2 }{\underline{\lambda} n} \right) \\
	& \leq C(d) \left( \kappa \widetilde{\sigma} \sqrt{\frac{ \sum_{k=1}^d p_k r_k + \prod_{k=1}^d r_k }{n}} +  \frac{ (\bar{r} \prod_{k=1}^d p_k )^{1/2} \kappa \widetilde{\sigma}^2 }{\underline{\lambda} n}  \right).
	\end{split}
\end{equation} 

Since the sample complexity in satisfying TRIP is smaller than the sample complexity needed in initialization \eqref{ineq: tensor regression two fact}, overall when $n \geq c(d) (\|\bcX^*\|_{\tHS}^2 + \sigma^2)  \frac{\kappa^2 \sqrt{\bar{r}} \bar{p}^{d/2} }{\underline{\lambda}^2} $ and $\bar{r} \leq \underline{p}^{1/2}$, we have the initialization condition in Theorem \ref{th: local contraction general setting} is satisfied. Then by Corollary \ref{coro: two-phase convergence} and the upper bound of $\|(\scA^*(\bvarepsilon))_{\max(2\br)}\|_{\tHS}$ in \eqref{ineq: tensor regression two fact}, when $$t_{\max} \geq C(d) \log \log \left( \frac{\underline{\lambda} \sqrt{n} }{\sigma\sqrt{\sum_{k=1}^d r_k p_k + \prod_{k=1}^d r_k } } \right),$$ we have
\begin{equation*}
		\|\bcX^{t_{\max}} - \bcX^*\|_{\tHS} \leq c(\sqrt{d} + 1)  \sigma\sqrt{\left(\sum_{k=1}^d r_k p_k + \prod_{k=1}^d r_k\right)/n}.
	\end{equation*} 
\quad $\blacksquare$

\subsection{Proof of Theorem \ref{th: tensor SVD} }
As we have mentioned in this setting, $\scA$ satisfies TRIP with $R_{2\br} =0, R_{3\br} = 0$ and here $\|(\scA^*(\bvarepsilon ))_{\max(2\br)}\|_{\tHS} = \|\bcE_{\max(2\br)}\|_{\tHS}$ where $\bcE$ is the noise in the tensorized model. Without loss of generality, we assume $\sigma = 1$.

First, we claim with probability at least $1-C\exp(-c\underline{p})$, the following inequalities hold:
\begin{equation} \label{ineq: tensor SVD two fact}
\begin{split}
	&\left\|\sin\Theta(\U_k^0,\U_k)\right\| = \|\U_{k\perp}^{0\top} \U_k\| \leq \frac{\sqrt{p_k} \underline{\lambda} + (\prod_{k=1}^d p_k)^{1/2}}{\underline{\lambda}^2} \\
	& \|\bcE_{\max(2\br)}\|_{\tHS} \leq C  \sqrt{\sum_{k=1}^d r_k p_k + \prod_{k=1}^d r_k }.
\end{split}
\end{equation} Here the first result holds when $\underline{\lambda} \geq c(d) \kappa \bar{p}^{d/4} \bar{r}^{1/4} $ by the proof of Theorem 1 in \cite{zhang2018tensor} and 
the second result is by Lemma \ref{lm: concentration of noise}.

Since $\bcX^0 = \bcY \times_{k=1}^d P_{\U_k^0}$, by Lemma \ref{lm: tensor estimation via projection}, we have
\begin{equation} \label{ineq: initial upper bound}
	\left\|\bcX^0 - \bcX^*\right\|_{\tHS} \leq \left\| \bcE \times_{k=1}^d P_{\U^0_k} \right\|_{\tHS} + \sum_{k=1}^d \left\| \U_{k\perp}^{0\top} \cM_k(\bcX^*) \right\|_F. 
\end{equation}

For $k=1,\ldots,d$
\begin{equation} \label{ineq: U0MkX}
	\begin{split}
		\left\| \U_k^{0\top} \cM_k(\bcX^*) \right\|_F =& \|\U_{k\perp}^{0\top} \U_k \U_k^\top \cM_k(\bcX^*)  \|_F\\
		\leq & \|\U_{k\perp}^{0\top} \U_k\|_F \|\U_k^\top \cM_k(\bcX^*)\|\\
		\overset{\eqref{ineq: tensor SVD two fact} } \leq & \frac{\sqrt{p_k r_k} \underline{\lambda} + ( r_k \prod_{k=1}^d p_k)^{1/2}}{\underline{\lambda}^2} \bar{\lambda}.
	\end{split}
\end{equation}
Notice $\left\| \bcE \times_{k=1}^d P_{\U^0_k} \right\|_{\tHS} \leq \|\bcE_{\max(2\br)}\|_{\tHS}$ by definition of $\|(\cdot)_{\max(2\br)}\|_{\tHS}$, combining \eqref{ineq: tensor SVD two fact} and \eqref{ineq: U0MkX}, from \eqref{ineq: initial upper bound}, we have
\begin{equation} \label{ineq: tensor-svd-initialization-guarantee}
\begin{split}
	\|\bcX^0 -\bcX^*\|_{\tHS} & \leq C(d) \left( \kappa \sum_{k=1}^d \sqrt{p_k r_k} + (\prod_{k=1}^d r_k)^{1/2} + \frac{ (\bar{r} \prod_{k=1}^d p_k )^{1/2} \kappa }{\underline{\lambda}} \right).
\end{split}
\end{equation}

Notice, when $\underline{\lambda} \geq c(d) \kappa \bar{p}^{d/4} \bar{r}^{1/4} $ and $\bar{r} \leq \underline{p}^{1/2}$, we have the initialization condition in Theorem \ref{th: local contraction general setting} is satisfied. Then by Corollary \ref{coro: two-phase convergence} and the upper bound of $\|\bcE_{\max(2\br)}\|_{\tHS}$ in \eqref{ineq: tensor SVD two fact}, when $$t_{\max} \geq C(d) \log \log \left( \frac{\underline{\lambda}  }{\sqrt{\sum_{k=1}^d r_k p_k + \prod_{k=1}^d r_k } } \right),$$ we have
\begin{equation*}
	\|\bcX^{t_{\max}} - \bcX^*\|_{\tHS} \leq 2(\sqrt{d} + 1) \sqrt{\sum_{k=1}^d r_k p_k + \prod_{k=1}^d r_k }.
\end{equation*} \quad $\blacksquare$

\section{Additional Proofs and Lemmas} \label{sec: additional lemmas}

{\noindent \bf Proof of Lemma \ref{lm: st-HOSVD as retraction}}. The proof is similar to the proof of \cite[Proposition 2.3]{kressner2014low}. Recall retraction $R$ is a smooth map from $T \bbM_{\br}$ to $\bbM_{\br}$ that satisfies 
\begin{itemize}
\item Property i) $R(\bcX, 0) = \bcX$;
\item Property ii) $\frac{d}{d t} R(\bcX, t \eta) \vert_{t = 0} = \eta$ for all $\bcX \in \bbM_{\br}$ and $\eta \in T_\bcX \bbM_{\br}$.
\end{itemize}

We begin by checking the smoothness of ST-HOSVD. For any tensor $\bcY$, let $\{\U_k^0\}_{k=1}^d$ be the mode-$k$ singular vectors computed in the ST-HOSVD algorithm. Let $D_1$ denotes the collection of tensors whose mode-$1$ matricization has a nonzero singular gap between the $r_1$th and $(r_1 + 1)$th singular values, and let $D_k$ be collection of tensors $\bcY$ such that $\cM_{r_k}(\bcY \times_{l=1}^{k-1} P_{\U_l^0} )$ has a nonzero singular gap between the $r_k$th and $(r_k + 1)$th singular values. Let $\mathcal{P}_{\U_k^0}$ be a projection operator in the tensor space such that $\mathcal{P}_{\U_k^0} \bcY = \bcY \times_k P_{\U_k^0}$. From standard results in matrix perturbation theory \citep{chern2001smoothness}, $\mathcal{P}_{\U_k^0}$ is smooth and well-defined on $D_k$. Since $\bcX$ is contained in all $D_k$ for $k=1,\ldots,d$ and is a fixed point of every $\mathcal{P}_{\U_k^0}$, we can construct an open neighborhood $D$ of $\bcX$ such that $\mathcal{P}_{\U_1^0} \circ \cdots \circ \mathcal{P}_{\U_d^0} D \subseteq D_k$ for all $k$ (here ``$\circ$'' is the composition of the projection operator). The smoothness of ST-HOSVD is implied by the chain rule.

Next, we check the two properties of retraction. Property i) is clear as $\bcX \in \bbM_{\br}$. For Property ii), because the tangent space $T_{\bcX} \bbM_{\br}$ is a first-order approximation of $\bbM_{\br}$ around $\eta$, we have $\|(\bcX + t\eta) - P_{\bbM_{\br}}(\bcX + t \eta)\| = O(t^2)$ as $t \to 0$. Thus, by the quasi-projection property of ST-HOSVD \citep[Chapter 10]{hackbusch2012tensor}, we have
\begin{equation*}
	\|(\bcX + t\eta) -{\text {\rm ST-HOSVD}}(\bcX + t \eta)\| \leq \sqrt{d} \|(\bcX + t\eta) -P_{\bbM_{\br}}(\bcX + t \eta)\| = O(t^2).
\end{equation*}
Hence, $R(\bcX + t\eta) = (\bcX + t \eta) + O(t^2)$ and $\frac{d}{d t} R(\bcX, t \eta) \vert_{t = 0} = \eta$. This has finished the proof.
\quad $\blacksquare$

{\noindent \bf Proof of Lemma \ref{lm:gradient}. } Since $\bbM_{\br}$ is an embedded submanifold of $\bbR^{p_1\times \cdots \times p_d}$, from \cite[(3.37)]{absil2009optimization}, we have the result. \quad $\blacksquare$

\begin{Lemma}[Tensor Restricted Orthogonal Property]\label{lm:retricted orthogonal property}
    Let $\bcZ_1, \bcZ_2 \in \bbR^{p_1 \times \cdots \times p_d}$ be two low Tucker rank tensors with $\tuckerrank(\bcZ_1) = \br_1 := (r_1,\ldots,r_d)$, $ \tuckerrank(\bcZ_2) = \br_2 := (r_1',\ldots,r_d')$.  If $\scA$ satisfies the $(\br_1+\br_2)$-TRIP condition, hen we have
    \begin{equation} \label{ineq: TROP-property1}
    	|\langle \scA(\bcZ_1), \scA(\bcZ_2) \rangle - \langle \bcZ_1, \bcZ_2 \rangle | \leq R_{\br_1 + \br_2} \|\bcZ_1\|_{\tHS} \|\bcZ_2\|_{\tHS}.
    \end{equation}
     In particular, if $\langle \bcZ_1, \bcZ_2 \rangle = 0$, we have
    \begin{equation*}
        |\langle \scA(\bcZ_1), \scA(\bcZ_2) \rangle | \leq R_{\br_1 + \br_2} \|\bcZ_1\|_{\tHS} \|\bcZ_2\|_{\tHS}.
    \end{equation*}
\end{Lemma}
{\noindent \bf Proof.} The proof for the matrix version of this result can be found in 
\cite[Lemma 3.3]{candes2011tight} and here we present the proof in the tensor setting. Without loss of generality, assume $\|\bcZ_1\|_{\tHS} = 1$, $\|\bcZ_2\|_{\tHS} = 1$. Notice that $\bcZ_1 + \bcZ_2$ is of at most Tucker rank $\br_1 + \br_2$ as the matricization of $\bcZ_1 + \bcZ_2$ on each mode $k$ is of at most rank $r_k + r_k'$. Similarly, $\bcZ_1 - \bcZ_2$ is also at most Tucker rank $\br_1 + \br_2$. By the TRIP of $\scA$, we have
\begin{equation*}
	\begin{split}
		& 2(1-R_{\br_1 + \br_2}) \pm 2(1-R_{\br_1 + \br_2}) \langle \bcZ_1, \bcZ_2 \rangle =(1-R_{\br_1 + \br_2}) \|\bcZ_1 \pm \bcZ_2 \|^2_{\tHS} \leq \| \scA(\bcZ_1 \pm \bcZ_2) \|_{\tHS}^2\\
		& 2(1+R_{\br_1 + \br_2}) \pm 2(1+R_{\br_1 + \br_2}) \langle \bcZ_1, \bcZ_2 \rangle =(1+R_{\br_1 + \br_2}) \|\bcZ_1 \pm \bcZ_2 \|^2_{\tHS} \geq \| \scA(\bcZ_1 \pm \bcZ_2) \|_{\tHS}^2.
	\end{split}
\end{equation*} 
Then we have
\begin{equation*}
	\begin{split}
		\langle \scA(\bcZ_1), \scA(\bcZ_2) \rangle  &= \frac{1}{4} \left(\| \scA(\bcZ_1 + \bcZ_2) \|_{\tHS}^2 - \scA(\bcZ_1 - \bcZ_2) \|_{\tHS}^2 \right) \leq R_{\br_1 + \br_2} +\langle \bcZ_1, \bcZ_2 \rangle  \\
		\langle \scA(\bcZ_1), \scA(\bcZ_2) \rangle  &= \frac{1}{4} \left(\| \scA(\bcZ_1 + \bcZ_2) \|_{\tHS}^2 - \scA(\bcZ_1 - \bcZ_2) \|_{\tHS}^2 \right) \geq -(R_{\br_1 + \br_2} +  \langle \bcZ_1, \bcZ_2 \rangle).
	\end{split}
\end{equation*}
Finally, we have
\begin{equation*}
	|\langle \scA(\bcZ_1), \scA(\bcZ_2) \rangle - \langle \bcZ_1, \bcZ_2 \rangle | \leq R_{\br_1 + \br_2}.
\end{equation*} \quad $\blacksquare$

\begin{Lemma}[Tensor Estimation from Projection] \label{lm: tensor estimation via projection}
	Given two order-$d$ tensors $\bcY, \bcX \in \bbR^{p_1 \times \cdots \times p_d}$. Suppose $\U_k^0 \in \bbO_{p_k,r_k}$, then
	\begin{equation*}
		\left\|\bcY \times_{k=1}^d P_{\U_k^0} - \bcX   \right\|_{\tHS} \leq \left\| (\bcY-\bcX) \times_{1} P_{\U^0_1} \times \cdots \times_{d} P_{\U^0_d}\right\|_{\tHS} + \sum_{k=1}^d \left\|\U_{k\perp}^{0\top} \mathcal{M}_{k}(\bcX)\right\|_F.
	\end{equation*} 
\end{Lemma}
{\noindent \bf Proof.}
 First, notice the following decomposition
\begin{equation*}
\begin{split}
\bcX = & \bcX \times_1 \Proj_{\U^0_{1}} \times \cdots \times_d P_{\U^0_{d}} + \bcX \times_1 P_{\U^0_{1\perp}} \times_2 P_{\U^0_{2}} \times \cdots \times_d P_{\U^0_{d}}\\
& + \bcX \times_1 \I_{p_{1}} \times_2 P_{\U^0_{2\perp}} \times \cdots \times_d P_{\U^0_{d}} + \bcX \times_1 \I_{p_{1}} \times_2 \I_{p_{2}} \times_3 P_{\U^0_{3\perp}}  \cdots \times_d P_{\U^0_{d}}\\
& + \ldots + \bcX \times^{d-1}_{i=1} \I_{p_{i}} \times_d P_{\U^0_{d\perp}}\\
= & \bcX \times_{1} \Proj_{\U^0_{1}} \times \cdots \times_{d} P_{\U^0_d} + \sum_{k=1}^d \bcX \times^{k-1}_{i =1} \I_{p_{i}} \times_{k} P_{\U^0_{k\perp}} \times^d_{i=k+1} P_{\U^0_{i}}.
\end{split}
\end{equation*}
Thus,
\begin{equation*}
\begin{split}
& \left\|\bcY \times_{1} P_{\U^0_1} \times \cdots \times_{d} P_{\U^0_d} - \bcX \right\|_{\tHS} \\
\leq & \left\|(\bcY-\bcX) \times_{1} P_{\U^0_1} \times \cdots \times_{d} P_{\U^0_d}\right\|_{\tHS} + \sum_{k=1}^d \left\|\bcX \times^{k-1}_{i= 1} \I_{p_{i}} \times_{k} P_{\U^0_{k\perp}} \times^d_{i=k+1} P_{\U^0_{i}}\right\|_{\tHS}\\
\leq & \left\| (\bcY-\bcX) \times_{1} P_{\U^0_1} \times \cdots \times_{d} P_{\U^0_d}\right\|_{\tHS} + \sum_{k=1}^d \left\|\U_{k\perp}^{0\top} \mathcal{M}_{k}(\bcX)\right\|_F.
\end{split}
\end{equation*}
\quad $\blacksquare$

\begin{Lemma}[Concentration of the Noise] \label{lm: concentration of noise}
	Suppose $\bcE \in \bbR^{p_1 \times \cdots \times p_d}$ and it has i.i.d. $N(0,1)$ entries. Then with probability at least $1 - \exp(-C\underline{p})$ ($\underline{p} := \min_{k=1,\ldots,d} p_k$), we have
	\begin{equation*}
		\left\|\bcE_{\max(2\br)}\right\|_{\tHS} \leq C'  \sqrt{\sum_{k=1}^d r_k p_k + \prod_{k=1}^d r_k },
	\end{equation*}
	for some $C' > 0$.
	
	 Suppose $\scA \in \mathbb{R}^{p_1\times \cdots \times p_d} \to \mathbb{R}^n$ is a linear map in \eqref{eq: affine operator} and its covariates $\bcA_i$ are independent and has i.i.d. $N(0,\frac{1}{n})$ entries, and $\bvarepsilon_i \overset{i.i.d.}\sim N(0,\frac{\sigma^2}{n} )$. Then with probability at least $1 - \exp(-C\underline{p})$, we have
	 \begin{equation*}
	 	\left\| (\scA^*(\bvarepsilon ))_{\max(2\br)} \right\|_{\tHS} \leq C'\sigma\sqrt{\frac{\sum_{k=1}^d r_k p_k + \prod_{k=1}^d r_k}{n} },
	 \end{equation*}
	 for some $C' > 0$.
\end{Lemma}
{\noindent \bf Proof.} The first result is proved in Lemma 5 of \cite{zhang2018tensor} in $d = 3$ case and it can be easily generalized to order-$d$ setting. 

For the second result, recall 
\begin{equation*}
\begin{split}
	\left\| (\scA^*(\bvarepsilon ))_{\max(2\br)} \right\|_{\tHS} = &\sup_{\U_k \in \bbO_{p_k,2r_k},k=1,\ldots,d} \|\scA^*(\bvarepsilon ) \times_{k=1}^d P_{\U_k} \|_{\tHS}\\
	=  &\sup_{\U_k \in \bbO_{p_k,2r_k},k=1,\ldots,d} \sup_{\bcZ: \|\bcZ\|_{\tHS} \leq 1} \langle \bcZ, \scA^*(\bvarepsilon ) \times_{k=1}^d P_{\U_k} \rangle\\
	= &\sup_{\U_k \in \bbO_{p_k,2r_k},k=1,\ldots,d} \sup_{\bcZ: \|\bcZ\|_{\tHS} \leq 1} \langle \bcZ \times_{k=1}^d P_{\U_k}, \scA^*(\bvarepsilon ) \rangle.
\end{split}
\end{equation*} It was proved in Theorem 4.2 of \cite{han2020optimal} that 
\begin{equation*}
	 \sup_{\U_k \in \bbO_{p_k,2r_k},k=1,\ldots,d} \sup_{\bcZ: \|\bcZ\|_{\tHS} \leq 1} \langle \bcZ \times_{k=1}^d P_{\U_k}, \scA^*(\bvarepsilon ) \rangle \leq 4 \sigma\sqrt{\frac{\sum_{k=1}^d r_k p_k + \prod_{k=1}^d r_k}{n} }
\end{equation*} 
holds with probability at least $1 - \exp(-C\underline{p})$ when $d = 3$. It is straightforward to extend to the order-$d$ case. For simplicity, we omit the proof here. \quad $\blacksquare$

\vskip 0.2in
\bibliography{reference}
\end{sloppypar}

\end{document}